\documentclass{article} % For LaTeX2e
\PassOptionsToPackage{sort&compress}{natbib}
\usepackage{arxiv,times}
\setcitestyle{numbers,square,comma}

\usepackage{amsmath,amsfonts,bm}

\def\eqref#1{equation~\ref{#1}}
\def\1{\bm{1}}

\DeclareMathAlphabet{\mathsfit}{\encodingdefault}{\sfdefault}{m}{sl}
\SetMathAlphabet{\mathsfit}{bold}{\encodingdefault}{\sfdefault}{bx}{n}

\usepackage{graphicx}
\usepackage{hyperref}
\usepackage{cleveref}
\usepackage{url}
\usepackage{xspace}
\usepackage{booktabs}
\usepackage{makecell}
\usepackage{multirow}
\usepackage{xcolor}
\usepackage{listings}
\usepackage{float}
\lstdefinestyle{prompt}{
  basicstyle=\ttfamily\footnotesize,breaklines=true,columns=fullflexible,
  frame=single,framesep=6pt
}
\lstdefinestyle{fa3algorithm}{
  basicstyle=\ttfamily\scriptsize,
  numbers=left,
  numberstyle=\tiny\color{black!55},
  stepnumber=1,
  numbersep=8pt,
  xleftmargin=2.5em,
  frame=single,
  framesep=5pt,
  breaklines=true,
  columns=fullflexible,
  keepspaces=true,
  mathescape=true,
  captionpos=b
}
\newcommand{\accell}[2]{%
  \makecell[c]{#2\\{\color{black!65}(#1)}}%
}

\newcommand{\spcell}[2]{%
  \makecell[c]{#1\\{\color{black!65}(#2)}}%
}
\newcommand{\method}{PTXBench\xspace}
\newcommand{\harness}{MiniPTXAgent\xspace}

\title{PTXBench: Evaluating and Adapting LLMs for GPU Kernel Optimization with Architecture-specific PTX}

\author{
Genghan Zhang\textsuperscript{1}\thanks{Correspondence to: Genghan Zhang \texttt{zgh23@stanford.edu}. Code available at \href{https://github.com/zhang677/PTXBench}{https://github.com/zhang677/PTXBench}.}
\quad
Yixin Dong\textsuperscript{2}
\quad
Chengze Fan\textsuperscript{4}
\quad
Zhichen Zeng\textsuperscript{3}
\quad
Yueming Yuan\textsuperscript{3}
\quad
Shaowei Zhu\textsuperscript{4}
\quad
Kunle Olukotun\textsuperscript{1}
\\[0.5em]
\textsuperscript{1}Stanford University
\quad
\textsuperscript{2}Carnegie Mellon University
\quad
\textsuperscript{3}RadixArk
\quad
\textsuperscript{4}Independent Researcher
}

\begin{document}

\maketitle

\begin{abstract}
We introduce PTXBench, a benchmark for evaluating and adapting large language models (LLMs) to use architecture-specific PTX for GPU kernel optimization. PTXBench measures functional correctness, whether selected target instructions execute at runtime, and speedup over frontier libraries across GEMM and attention workloads on H100 and B200 GPUs. Our evaluation shows that architecture-specific PTX capability remains uneven: success rates fall substantially on complex attention backward workloads, and executing the target instructions does not necessarily translate into competitive performance. No evaluated model consistently matches frontier libraries across the suite. We further adapt Qwen3.6-27B using supervised fine-tuning. Repair-conditioned training improves several tasks, but generalization remains uneven; data coverage, balance, and the quality of the reasoning teacher matter in addition to dataset size. PTXBench provides an auditable testbed for measuring and improving LLMs' ability to exploit evolving GPU architectures.
\end{abstract}

\section{Introduction}
PTX (Parallel Thread Execution) is the lowest-level programmable interface that CUDA developers can explicitly control. High-performance kernels often require PTX-level optimization to efficiently leverage architecture-specific mechanisms~\citep{deepgemm2025}. Recent NVIDIA GPU generations have been introducing new architecture-specific PTX instructions for tensor cores, memory units, and synchronization mechanisms~\citep{nvidia-volta-tuning,nvidia-ampere-tuning,
nvidia-hopper-tuning,nvidia-blackwell-tuning,nvidia-rubin-architecture}. A portable CUDA kernel can remain functionally correct while leaving the defining capabilities of newer hardware unused. Higher-level kernel languages aim to recover performance portability~\citep{tillet2019triton,cutlass2023}, which ultimately lower to the same PTX interface. However, keeping these kernel languages aligned with rapidly evolving hardware requires continued compiler engineering~\citep{chen2026tawa}, while validating an optimizing compiler entails a much larger input and configuration space and thus is much harder than validating an individual kernel~\citep{dubey2025equivalence}. This raises a practical question for kernel developers: can an LLM directly exploit architecture-specific features, and can targeted post-training improve this ability?

To answer this question, we introduce \method, an architecture-specific GPU kernel benchmark that evaluates whether LLMs can exploit specified low-level GPU mechanisms. \method incorporates \harness, a multi-turn agent loop in which models are given an accurate architecture-specific knowledge pack and generate CUDA kernels with inline PTX (which we call CUDA--PTX) and revise them using structured execution feedback. Across each trajectory, \method separately measures functional correctness, whether the required instruction family executes at runtime under the evaluated workload, and peak and typical performance. \method is built on FlashInfer-Trace~\citep{xing2026flashinfer}, a unified schema for kernel workloads, solutions, and evaluations. Separating the capability probe from the problem collection makes \method extensible to suites with compatible interfaces, such as SOL-ExecBench~\citep{lin2026sol}. Existing GPU kernel benchmarks, beginning with KernelBench~\citep{ouyang2025kernelbench}, ask models to replace reference PyTorch operators with functionally correct, faster GPU kernels~\citep{saroufim2025backendbench,li2025tritonbench}. These end-to-end outcomes are essential, but they do not isolate whether a model can directly program a specified architecture mechanism: performance may instead come from generic CUDA code or calls to existing vendor libraries. \method complements these suites with a controlled capability probe: models must directly produce efficient low-level kernels using a specified family of architecture-specific PTX instructions, and the corresponding target instructions must execute at runtime.

\method provides a controlled measurement framework and an environment for collecting adaptation data. We first characterize architecture-specific PTX capability across current models and GPUs, and then test targeted, repair-conditioned post-training. 

This work makes three contributions:
\begin{itemize}
    \item \textbf{An architecture-specific PTX benchmark.} We introduce \method, a multi-turn benchmark for evaluating whether LLMs can produce functionally correct GPU kernels that execute the required target instructions at runtime. It provides controlled architecture knowledge and separately measures correctness, target instruction execution, and performance relative to frontier libraries.
    
    \item \textbf{A capability study across models, architectures, and workloads.} We evaluate closed- and open-weight models on H100 and B200 across GEMM and attention workloads. Models frequently succeed on forward workloads but struggle with backward attention, and even kernels with verified target instruction execution generally remain slower than frontier libraries. These results expose a substantial gap between executing architecture-specific instructions, producing correct kernels, and achieving competitive performance.
    
    \item \textbf{A controlled adaptation study.} We conduct, to our knowledge, the first controlled study of SFT conditioned on repairs for CUDA and PTX generation that targets a specific architecture, adapting Qwen3.6-27B and ablating problem coverage and balance and reasoning supervision. Supervised fine tuning conditioned on repairs improves several tasks, but transfer to held-out shapes and attention variants remains uneven. Problem coverage, data balance, and the reasoning teacher all matter.
\end{itemize}

\begin{figure*}[htbp]
  \centering
  \includegraphics[width=\textwidth]{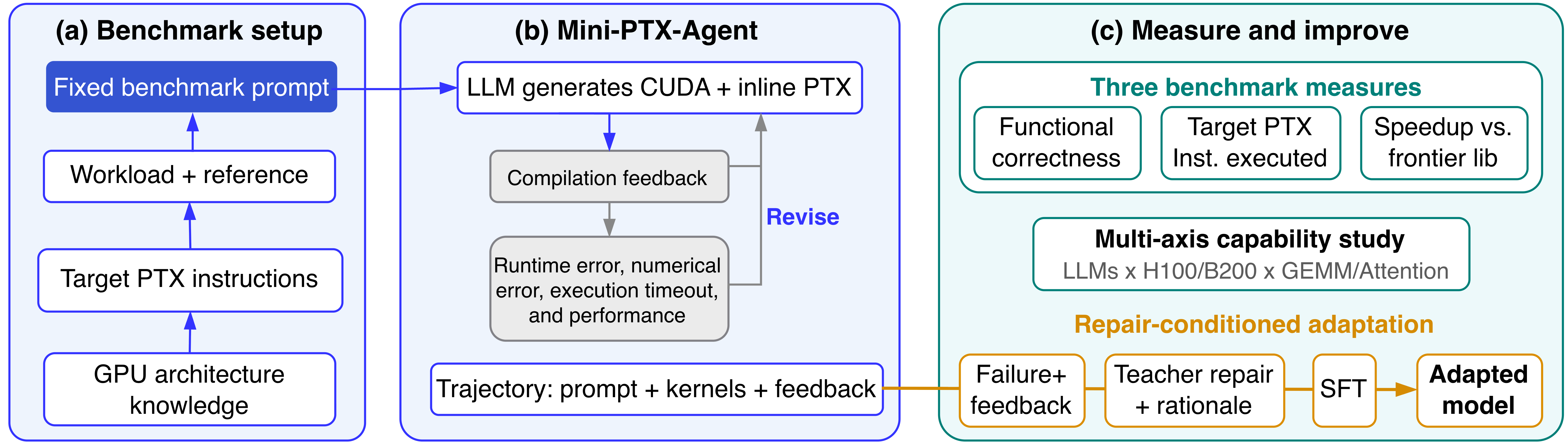}
  \caption{Overview of the \method benchmark and adaptation workflow.}
  \label{fig:ptxbench-overview}
\end{figure*}

\section{\method}
\Cref{fig:ptxbench-overview} summarizes \method's benchmark and adaptation workflow. \method is designed around three requirements for evaluating architecture-specific PTX programming. First, models receive controlled architecture knowledge because they otherwise rarely use the requested PTX. Second, we verify that the required instruction family executes at runtime under the fixed workload. Third, we evaluate each kernel for correctness and efficiency relative to frontier library implementations.

\subsection{Benchmark Tasks and Controlled Context}

A \method instance pairs a reference operator composed from frontier libraries such as cuBLAS, cuDNN, and FlashInfer~\citep{nvidia_cublas,chetlur2014cudnn,ye2025flashinfer} with a fixed workload, target GPU architecture, and required family of architecture-specific PTX instructions. The model generates a CUDA kernel with inline PTX from scratch rather than starting from an initial implementation. \method uses the FlashInfer-Trace schema to express workloads, solutions, and correctness checks, allowing the same benchmark workflow to support other compatible task collections.

To measure architecture-specific reasoning rather than documentation retrieval, every trajectory receives the same architecture-specific knowledge pack in its base prompt: architecture parameters, CUDA wrappers for PTX instructions, and contracts governing layouts, synchronization, and memory consistency. For well-studied operators, we also provide fixed, expert-validated scheduling principles (\Cref{app:fa3-algorithm} shows an example for FlashAttention). Each pack occupies 20k--30k tokens (\Cref{app:prompt-token-counts}). The ablation in~\Cref{tab:gemini31-knowledge-ablation} shows that without this context, the model rarely uses the requested PTX.

\subsection{Multi-turn Evaluation Protocol}

\harness uses a multi-turn protocol because iterative kernel generation, execution feedback, and revision form the fundamental operating loop of kernel agents~\citep{cao2026ksearch,zhang2026accelopt,hong2025autocomp}. Each trajectory permits a fixed number of model calls and retains all preceding kernels and execution feedback. \harness first compiles each candidate with \texttt{nvcc} in a CPU container, sending only successful compilations to a profiling service for memory-safety checking, runtime error messages, functional evaluation, and latency measurement.

We define \textbf{target instruction correctness} as functional correctness plus execution of selected tensor-compute or data-movement instructions at runtime under the evaluated workload. To measure execution, we analyze SASS, NVIDIA's native GPU assembly generated from PTX and executed by the GPU. For each functionally correct candidate, NVIDIA Nsight Compute (NCU) supplies predicate-enabled thread execution counts for the instruction families in~\Cref{tab:architecture-sass-detection}. A candidate qualifies when either the architecture-specific compute predicate $C$ or data-movement predicate $D$ is true. On B200, TMA payload transfer can satisfy $D$, while fifth-generation tensor compute requires both MMA and TMEM readback. Static SASS presence alone does not establish execution; failed profiling remains unknown. Target instruction correctness establishes whether the selected instructions execute, not how much useful work they perform or whether they cause the measured speedup. \Cref{app:target-inst-measurement} gives implementation details and edge cases.

The functional correctness checker verifies output shape and data type, then compares values using \texttt{torch.allclose} with \texttt{atol=rtol=1e-2}. After a candidate passes the correctness check, candidate and reference latencies are measured with CUPTI~\citep{nvidia2026cupti}. For each trial, we take the median over 50 timed iterations after 10 warmup iterations, and compute speedup as reference latency divided by candidate latency. Including header files from cuDNN or cuBLAS~\citep{nvidia_cublas} in CUDA source code is considered wrong regardless of execution results. Generated kernels may crash, hang, or corrupt subsequent measurements, so the profiling service isolates execution of kernels from the agent and exposes correctness, sanitization, debugging, and latency measurement as independent operations. More details on the profiling service are in~\Cref{app:system-efficiency}.

\section{Adapting LLMs to Architecture-Specific PTX Programming}
\label{sec:adaptation-method}
Deployed GPUs retain fixed low-level capabilities for years, making architecture-specific PTX a durable target for post-training. The obstacle is data: high-quality kernels require scarce expert knowledge and iterative validation. We therefore study whether model failures and execution feedback, paired with teacher-generated repairs and rationales, can provide effective supervision for this domain.

\paragraph{Fixit.}
Fixit constructs supervision from failures produced by the model being adapted. Let $x$ denote a problem prompt together with its architecture-specific context, $\pi_0$ the model before adaptation, $H$ the \harness feedback function, and $C$ the functional-correctness indicator. We first sample a failed kernel $k^-$ and collect its compilation or execution feedback $e$:
\[
k^- \sim \pi_0(\cdot \mid x), \qquad e = H(x,k^-), \qquad C(x,k^-)=0.
\]
A repair teacher $\pi_R$ then generates a corrected kernel conditioned on the same problem, failure, and feedback. We retain only repairs $k^+$ that pass the correctness check:
\[
k^+ \sim \pi_R(\cdot \mid x,k^-,e), \qquad C(x,k^+)=1.
\]
Finally, a reasoning teacher $\pi_T$ synthesizes a rationale $r$ that leads from the observed failure to the retained repair:
\[
r \sim \pi_T(\cdot \mid x,k^-,e,k^+).
\]
Each Fixit example therefore conditions the student on $(x,k^-,e)$ and supervises it with $(r,k^+)$. Failures are collected once from the model before adaptation, so the supervision targets error states produced by that fixed checkpoint; the repairs and rationales are teacher-generated.

\section{Benchmark Results}

\subsection{Evaluation Metrics and Baselines}
We evaluate models along four complementary dimensions. Turn correctness rate (\# of correct kernels / \# of turns) captures the ability to generate correct kernels. Target instruction turn correctness rate (\# of correct kernels that execute a selected target instruction / \# of turns) evaluates whether a model can produce a functionally correct kernel that exercises the requested architecture mechanism. Finally, among correct turns, best speedup measures peak optimization ability, while target instruction best speedup restricts that peak to qualifying kernels. Inspired by KernelBench~\citep{ouyang2025kernelbench}, for $N$ evaluated turns with correctness $C_i$, runtime target instruction indicator $I_i$, and speedup $s_i$, we summarize the full speedup distribution as
\[
\mathrm{Fast}_p=\frac{1}{N}\sum_{i=1}^{N}\mathbf{1}[C_i=1\land s_i>p],\qquad
\mathrm{Fast}^{\mathrm{Inst.}}_p=\frac{1}{N}\sum_{i=1}^{N}\mathbf{1}[C_i=1\land I_i=1\land s_i>p].
\]
At threshold $p$, $\mathrm{Fast}^{\mathrm{Inst.}}_p$ is the fraction of all turns that produce a correct kernel, execute a selected target instruction at runtime, and exceed speedup $p$. NCU failures remain unknown and do not enter the target instruction numerator, so the reported values are conservative. In space-constrained table headers and plot axes, we abbreviate target instruction as ``Target inst.'' Across all result tables, the first line is the target instruction metric, the parenthesized line is unrestricted, and triplets report $\leq1/\leq4/\leq8$ turns. Plot curves show the corresponding turn fractions, and vertical labels mark the best qualifying speedup. Solid lines show the mean across three prompt variants, each evaluated over four eight-turn trajectories ($N=32$), while shading spans the prompt-wise minimum and maximum. Throughout the paper, Fwd denotes multihead attention (MHA) forward with LSE input, Bwd denotes MHA backward, Causal denotes causal attention, and \texttt{d} denotes the head dimension. For GQA variants, the performance baseline is FlashInfer (v0.6.14); for all other attentions, it is cuDNN (v9.20.0); for GEMM, it is cuBLAS (v13.1.0). These baselines represent frontier performance, and achieving such performance on these problems requires complex scheduling of PTX instructions specific to each architecture.

\begin{table*}[htbp]
\centering
\scriptsize
\setlength{\tabcolsep}{2pt}
\renewcommand{\arraystretch}{1.12}
\caption{Target-instruction and unrestricted turn correctness rates (\%) on H100 and B200.}
\label{tab:turn-correctness}
\begin{tabular*}{\textwidth}{@{\extracolsep{\fill}}llccccc@{}}
\toprule
GPU
& Model
& GEMM
& MHA-Fwd
& MHA-Fwd-Causal
& MHA-Bwd
& MHA-Bwd-Causal \\
\midrule
\multirow{5}{*}{H100}
&
Gemini 3.1 Pro
& \accell{33.3 / 62.5 / 59.4}{33.3 / 60.4 / 56.2}
& \accell{33.3 / 39.6 / 45.8}{33.3 / 39.6 / 45.8}
& \accell{25.0 / 52.1 / 55.2}{25.0 / 52.1 / 55.2}
& \accell{8.3 / 33.3 / 38.5}{8.3 / 33.3 / 38.5}
& \accell{8.3 / 25.0 / 26.0}{-- / 22.9 / 25.0}
\\

&
Claude Opus 4.8
& \accell{91.7 / 95.8 / 94.8}{91.7 / 95.8 / \textbf{94.8}}
& \accell{66.7 / 89.6 / 94.8}{50.0 / 81.2 / \textbf{90.6}}
& \accell{83.3 / 89.6 / 82.3}{50.0 / 77.1 / \textbf{75.0}}
& \accell{50.0 / 83.3 / 89.6}{8.3 / 62.5 / \textbf{79.2}}
& \accell{25.0 / 41.7 / 59.4}{-- / 22.9 / 44.8}
\\

&
GLM-5.2
& \accell{33.3 / 62.5 / 62.5}{33.3 / 60.4 / 61.5}
& \accell{16.7 / 14.6 / 15.6}{16.7 / 8.3 / 8.3}
& \accell{8.3 / 16.7 / 17.7}{-- / 8.3 / 10.4}
& \accell{8.3 / 18.8 / 16.7}{8.3 / 6.2 / 5.2}
& \accell{-- / 22.9 / 15.6}{--}
\\

&
Qwen3.6-27B
& --
& --
& --
& --
& --
\\
&
GPT-5.6 xhigh
& \accell{91.7 / 83.3 / 79.2}{91.7 / 81.2 / 77.1}
& \accell{16.7 / 58.3 / 69.8}{8.3 / 39.6 / 60.4}
& \accell{-- / 64.6 / 63.5}{-- / 50.0 / 56.2}
& \accell{58.3 / 60.4 / 65.6}{8.3 / 39.6 / 55.2}
& \accell{33.3 / 56.2 / 65.6}{-- / 31.2 / \textbf{52.1}}
\\
\midrule
\multirow{5}{*}{B200}
& Gemini 3.1 Pro
& \accell{8.3 / 47.9 / 45.8}{8.3 / 47.9 / 45.8}
& \accell{-- / 12.5 / 15.6}{-- / 12.5 / 15.6}
& \accell{8.3 / 12.5 / 12.5}{-- / 8.3 / 10.4}
& \accell{-- / 2.1 / 2.1}{-- / 2.1 / 2.1}
& \accell{-- / 2.1 / 1.0}{--} \\

&
Claude Opus 4.8
& \accell{75.0 / 81.2 / 88.5}{25.0 / 64.6 / \textbf{80.2}}
& \accell{83.3 / 87.5 / 86.5}{-- / 20.8 / 38.5}
& \accell{91.7 / 77.1 / 83.3}{-- / 16.7 / 32.3}
& \accell{83.3 / 89.6 / 91.7}{-- / 6.2 / 10.4}
& \accell{25.0 / 52.1 / 68.8}{--} \\

&
GLM-5.2
& \accell{33.3 / 37.5 / 34.4}{16.7 / 27.1 / 28.1}
& \accell{33.3 / 22.9 / 26.0}{-- / -- / 1.0}
& \accell{33.3 / 22.9 / 25.0}{8.3 / 2.1 / 1.0}
& \accell{8.3 / 25.0 / 30.2}{--}
& \accell{16.7 / 18.8 / 22.9}{--} \\

&
Qwen3.6-27B
& \accell{-- / -- / 1.0}{--}
& --
& --
& --
& -- \\
&
GPT-5.6 xhigh
& \accell{50.0 / 66.7 / 69.8}{-- / 45.8 / 59.4}
& \accell{16.7 / 43.8 / 52.1}{-- / 18.8 / \textbf{39.6}}
& \accell{8.3 / 43.8 / 55.2}{-- / 18.8 / \textbf{42.7}}
& \accell{75.0 / 54.2 / 54.2}{-- / 22.9 / \textbf{37.5}}
& \accell{83.3 / 47.9 / 47.9}{-- / 22.9 / \textbf{31.2}}
\\
\bottomrule
\end{tabular*}
\end{table*}

\newcommand{\modelspeeduptable}{%
\begin{table*}[htbp]
\centering
\scriptsize
\setlength{\tabcolsep}{1pt}
\renewcommand{\arraystretch}{1.12}
\caption{Best speedup with target instruction execution on H100 and B200.}
\label{tab:speedup}
\resizebox{\textwidth}{!}{%
\begin{tabular*}{1.025\textwidth}{@{\extracolsep{\fill}}l@{\hspace{4pt}}lccccc@{}}
\toprule
GPU
& Model
& GEMM
& MHA-Fwd
& MHA-Fwd-Causal
& MHA-Bwd
& MHA-Bwd-Causal
\\
\midrule
\multirow{5}{*}{H100}
&
Gemini 3.1 Pro
& \spcell{0.687 / 0.934 / 0.962}{0.687 / 0.934 / 0.962}
& \spcell{0.555 / 0.730 / 0.730}{0.555 / 0.730 / 0.730}
& \spcell{0.614 / 0.651 / 0.768}{0.614 / 0.651 / 0.768}
& \spcell{0.206 / 0.375 / 0.515}{0.206 / 0.375 / 0.515}
& \spcell{-- / 0.634 / 0.639}{0.065 / 0.634 / 0.639}
\\

&
Claude Opus 4.8
& \spcell{0.770 / 0.968 / \textbf{0.976}}{0.770 / 0.968 / 0.976}
& \spcell{0.759 / 0.770 / 0.839}{0.759 / 0.770 / 0.839}
& \spcell{0.758 / 0.806 / 0.806}{0.758 / 0.806 / 0.806}
& \spcell{0.300 / 0.440 / 0.489}{0.300 / 0.440 / 0.489}
& \spcell{-- / 0.499 / 0.499}{0.058 / 0.499 / 0.499}
\\

&
GLM-5.2
& \spcell{0.447 / 0.692 / 0.692}{0.447 / 0.692 / 0.692}
& \spcell{0.407 / 0.470 / 0.607}{0.407 / 0.470 / 0.607}
& \spcell{-- / 0.471 / 0.533}{0.015 / 0.471 / 0.533}
& \spcell{0.316 / 0.437 / 0.437}{0.316 / 0.437 / 0.437}
& \spcell{--}{-- / 0.101 / 0.101}
\\

&
Qwen3.6-27B
& --
& --
& --
& --
& --
\\
&
GPT-5.6 xhigh
& \spcell{0.788 / 0.870 / 0.870}{0.788 / 0.870 / 0.870}
& \spcell{0.282 / 0.774 / \textbf{0.872}}{0.282 / 0.774 / 0.872}
& \spcell{-- / 0.699 / \textbf{0.865}}{-- / 0.699 / 0.865}
& \spcell{0.230 / 0.699 / \textbf{0.703}}{0.230 / 0.699 / 0.703}
& \spcell{-- / 0.650 / \textbf{0.746}}{0.094 / 0.650 / 0.746}
\\
\midrule
\multirow{5}{*}{B200}
& Gemini 3.1 Pro
& \spcell{0.273 / 0.680 / 0.892}{0.273 / 0.680 / 0.892}
& \spcell{-- / 0.280 / 0.280}{-- / 0.280 / 0.280}
& \spcell{-- / 0.206 / 0.248}{0.013 / 0.206 / 0.248}
& \spcell{-- / 0.087 / 0.133}{-- / 0.087 / 0.133}
& \spcell{--}{-- / 0.015 / 0.015} \\

&
Claude Opus 4.8
& \spcell{0.782 / 1.012 / \textbf{1.012}}{0.782 / 1.012 / 1.012}
& \spcell{-- / 0.253 / 0.300}{0.110 / 0.253 / 0.300}
& \spcell{-- / 0.232 / 0.269}{0.042 / 0.232 / 0.269}
& \spcell{-- / 0.069 / 0.149}{0.022 / 0.136 / 0.155}
& \spcell{--}{0.023 / 0.088 / 0.116} \\

&
GLM-5.2
& \spcell{0.162 / 0.632 / 0.632}{0.162 / 0.632 / 0.632}
& \spcell{-- / -- / 0.027}{0.024 / 0.024 / 0.035}
& \spcell{0.098 / 0.098 / 0.098}{0.098 / 0.098 / 0.098}
& \spcell{--}{0.019 / 0.030 / 0.040}
& \spcell{--}{0.018 / 0.034 / 0.034} \\

&
Qwen3.6-27B
& \spcell{--}{-- / -- / 0.006}
& --
& --
& --
& -- \\
&
GPT-5.6 xhigh
& \spcell{-- / 0.781 / 0.818}{0.175 / 0.781 / 0.818}
& \spcell{-- / 0.419 / \textbf{0.536}}{0.068 / 0.419 / 0.536}
& \spcell{-- / 0.294 / \textbf{0.445}}{0.132 / 0.294 / 0.445}
& \spcell{-- / 0.361 / \textbf{0.372}}{0.059 / 0.361 / 0.372}
& \spcell{-- / 0.339 / \textbf{0.339}}{0.066 / 0.339 / 0.339}
\\
\bottomrule
\end{tabular*}
}
\end{table*}
}

\begin{figure*}[t]
  \centering
  \includegraphics[width=\textwidth]{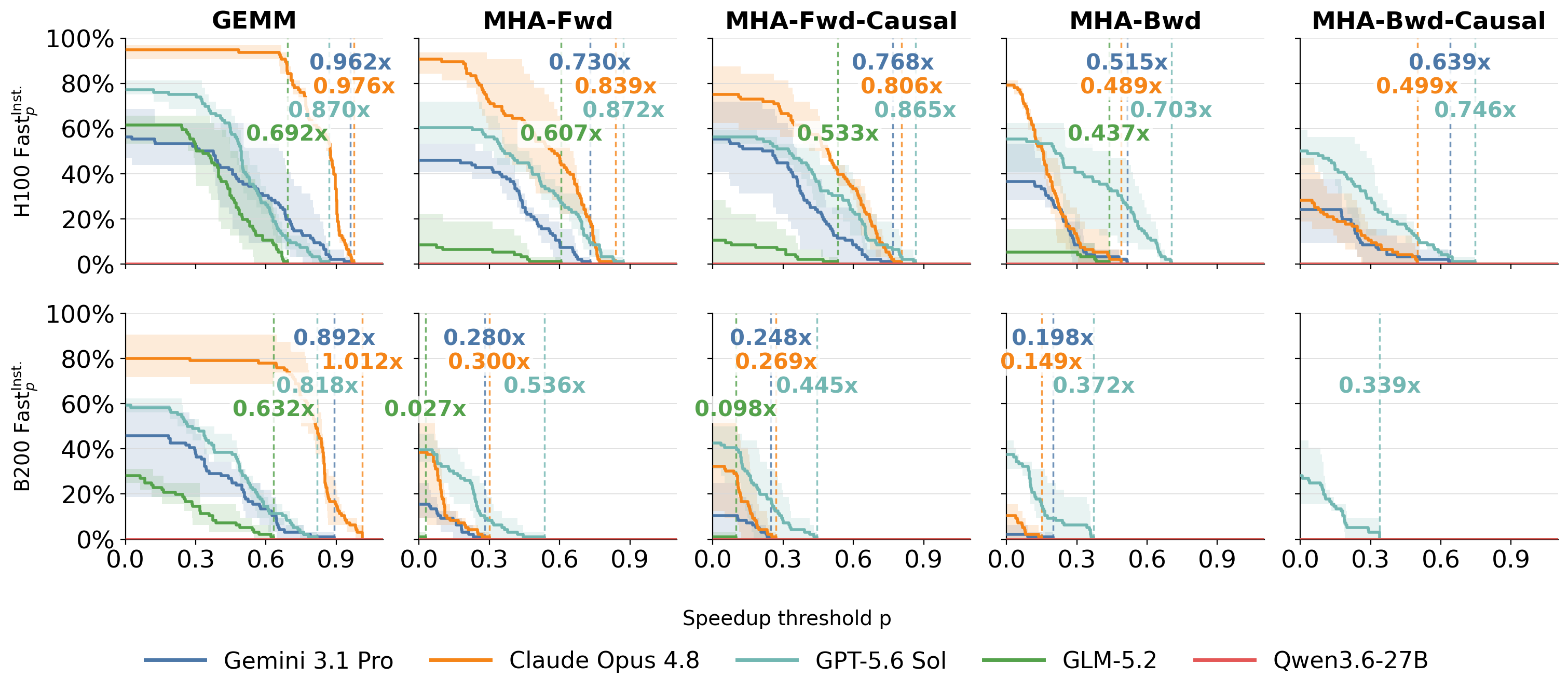}
  \caption{$\mathrm{Fast}^{\mathrm{Inst.}}_p$ on H100 (top) and B200 (bottom).}
  \label{fig:target-ptx-fast-at-p}
\end{figure*}

\subsection{Architecture-Specific PTX Capability Remains Uneven}
\Cref{tab:turn-correctness,fig:target-ptx-fast-at-p} compare models on H100 and B200. The $\mathrm{Fast}^{\mathrm{Inst.}}_p$ curves show both how often each model produces qualifying kernels and the distribution of their speedups. Since workload difficulty varies, we report speedups per problem; Appendix~\Cref{tab:speedup} gives exact values.

\begin{table}[htbp]
    \centering
    \caption{Model release dates, knowledge cutoffs, vendor-reported SWE-bench Pro scores, and estimated calendar
    lag from the release of Hopper PTX ISA 8.0 (Dec. 2022) and Blackwell PTX ISA 8.7 (Jan. 2025)~\citep{nvidia2022ptx80,nvidia2025ptx87}. Lags use monthly granularity, from PTX release to disclosed cutoff; otherwise, model release dates provide upper bounds.}
    \label{tab:model-ptx-cutoffs}
    \small
    \setlength{\tabcolsep}{4pt}
    \begin{tabular}{lccccc}
        \toprule
        \textbf{Model}
        & \makecell{\textbf{SWE-bench}\\\textbf{Pro (\%)}}
        & \textbf{Model release}
        & \textbf{Knowledge cutoff}
        & \multicolumn{2}{c}{\textbf{Lag after PTX release (months)}} \\
        \cmidrule(lr){5-6}
        & & & & \textbf{Hopper} & \textbf{Blackwell} \\
        \midrule

        Gemini 3.1 Pro
        & 54.2 & Feb.\ 2026 & Jan.\ 2025 & 25 & $\approx 0$ \\

        Claude Opus 4.8
        & 69.2 & May 2026 & Jan.\ 2026 & 37 & 12 \\

        GPT-5.6 Sol
        & 64.6 & July 2026 & Feb.\ 2026 & 38 & 13 \\

        GLM-5.2
        & 62.1 & June 2026 & Not disclosed & $\leq 42$ & $\leq 17$ \\

        Qwen3.6-27B
        & 53.5 & Apr.\ 2026 & Not disclosed & $\leq 40$ & $\leq 15$ \\

        \bottomrule
    \end{tabular}
\end{table}

We choose these five models because they were released relatively close together in time~\citep{google2026gemini31pro,anthropic2026opus48,openai2026gpt56,zai2026glm52,qwen2026qwen36}. \Cref{tab:model-ptx-cutoffs} summarizes their public release metadata, vendor-reported SWE-bench Pro scores, and intervals during which the PTX documentation could have entered the training data.

\noindent\begin{minipage}{\linewidth}
The SWE-bench Pro results provide context for general coding ability, but they are not a controlled comparison because agent scaffolds and evaluation settings differ. They also do not fully characterize architecture-specific capability: Gemini 3.1 Pro and Qwen3.6-27B differ by only 0.7 points on SWE-bench Pro, yet Gemini executes selected Hopper instructions across all five workloads, whereas Qwen3.6-27B produces no correct Hopper kernel under our setup.
\end{minipage}

Calendar exposure alone is similarly incomplete. Although Gemini 3.1 Pro's knowledge cutoff is in the same month as the Blackwell PTX release, it still achieves $0.892\times$ cuBLAS performance on GEMM on Blackwell. Claude Opus 4.8 achieves a substantially higher target instruction correctness rate on Blackwell and reaches $1.012\times$ cuBLAS on GEMM, but neither model optimizes Blackwell attention as well as Hopper attention. Together, these results suggest that general coding ability and newer PTX knowledge can help, but neither public coding scores nor calendar exposure alone predicts architecture-specific PTX capability or performance.

\begin{figure*}[t]
  \centering
  \includegraphics[width=\textwidth]{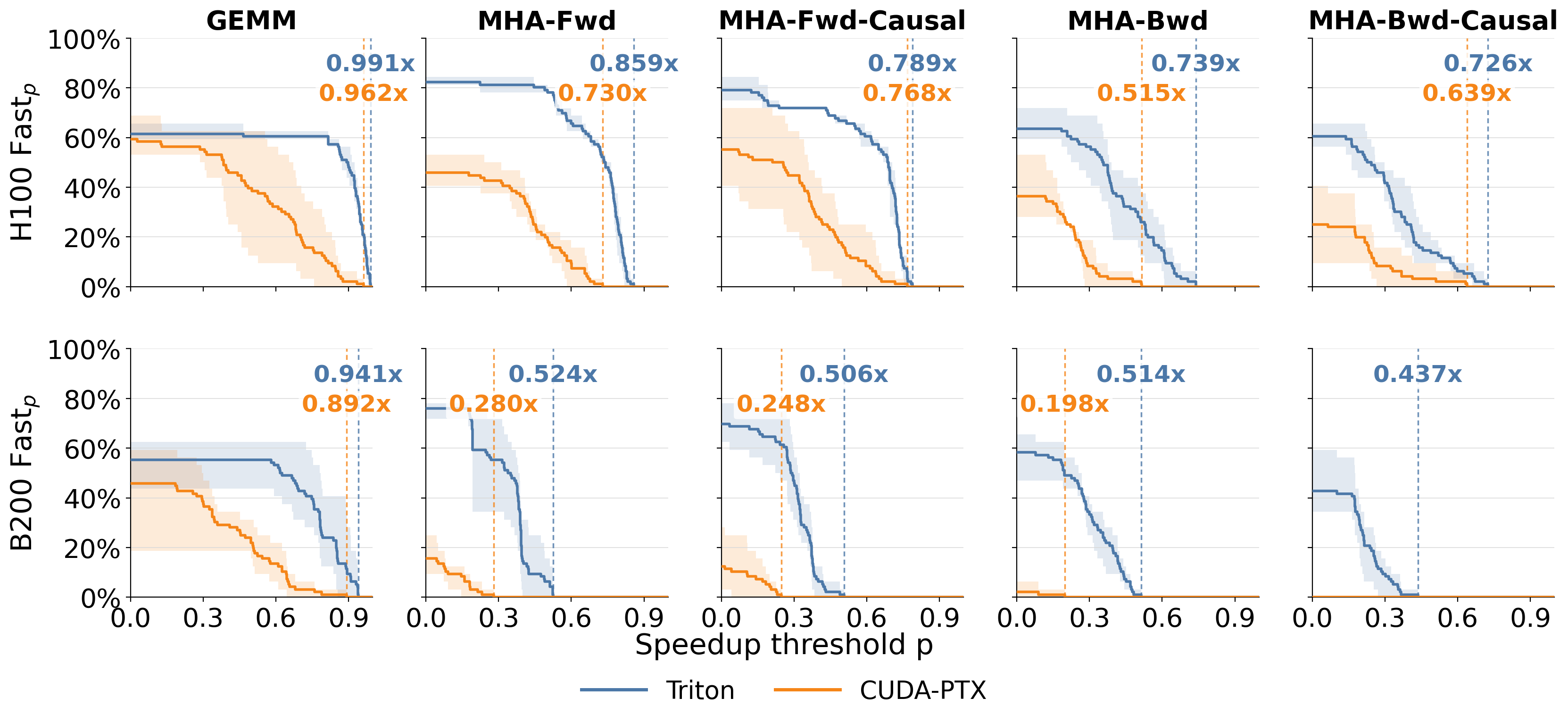}
  \caption{Gemini 3.1 Pro $\mathrm{Fast}_p$ distributions for Triton and CUDA-PTX on H100 and B200.}
  \label{fig:triton-cuda-fast-at-p}
\end{figure*}

Among models with open weights, GLM-5.2 is competitive with Gemini 3.1 Pro for GEMM on both H100 and B200 and for attention workloads without causal masking on H100, but it still lags on B200 attention workloads. Like Claude Opus 4.8, GLM-5.2 also tends to fall back on generic CUDA instead of architecture-specific PTX when the target is the newer Blackwell architecture. In contrast, Gemini 3.1 Pro is more successful at executing Blackwell instructions, achieving similar peak performance to Claude Opus 4.8 despite its lower success rate. Qwen3.6-27B produces one correct GEMM kernel on Blackwell, but it does not execute any selected Blackwell instruction.

\subsection{High-Level Kernel Languages Remain Valuable on New Architectures}

\Cref{fig:triton-cuda-fast-at-p} compares Gemini 3.1 Pro kernels generated in Triton and CUDA--PTX using the same five workloads on each GPU. On H100, the peak speedups are close on GEMM ($0.991\times$ for Triton versus $0.962\times$ for CUDA--PTX), but Triton has higher $\mathrm{Fast}_p$ across the attention workloads. On B200, the attention gap is larger: Triton reaches $0.514\times$ on MHA-Bwd and $0.437\times$ on MHA-Bwd-Causal, compared with $0.198\times$ and no correct CUDA--PTX candidate, respectively. The corresponding GPT-5.6 Sol comparison appears in \Cref{fig:gpt56-triton-cuda-fast-at-p}, including cases where CUDA--PTX attains a higher single-kernel peak.

Two factors may explain this architecture-dependent gap. First, Blackwell adds more specialized compute and memory mechanisms, increasing the complexity of coordinating low-level instructions directly. Second, Blackwell is roughly two years newer than Hopper, so far less Blackwell-specific CUDA-PTX code may have appeared in model training data (cf.\ \Cref{tab:model-ptx-cutoffs}). We also carefully prepared architecture-specific prompts for CuTeDSL, but its kernel success rate remained extremely low, preventing a meaningful performance comparison.

The broader takeaway is that high-level kernel languages remain useful for producing correct kernels on new architectures, where directly generating CUDA-PTX can be less effective. They do not, however, guarantee the best attainable performance: once correct, direct low-level implementations can sometimes match or outperform a higher-level implementation.

\subsection{Comparison with Coding Agents}
\label{sec:antigravity-comparison}
We compare Antigravity, using Gemini 3.1 Pro, with multi-turn refinement using the same model and task specification on H100. Each method has four trajectories per workload and eight evaluator requests per trajectory. Both receive the same architectural knowledge and execution feedback without scheduling guidance. Antigravity can edit files and call the same profiling service through an evaluation client; model calls and tool use can differ between harnesses. We therefore compare outcomes at a fixed evaluation-request budget and report token cost separately. Here the outcome is each trajectory's best correct, runtime-instruction-qualified speedup, or zero if none qualifies; \Cref{tab:agy-agent-comparison} reports the median over four trajectories.

\begin{table*}[htbp]
\centering
\small
\setlength{\tabcolsep}{4pt}
\caption{Antigravity versus multi-turn refinement after eight H100 evaluation requests per trajectory. Speedups are medians of four per-trajectory best instruction-qualified values. Token ratios divide the agent's median cumulative total tokens by the refinement median within each workload.}
\label{tab:agy-agent-comparison}
\begin{tabular*}{\textwidth}{@{\extracolsep{\fill}}lccccc@{}}
\toprule
Method or ratio & GEMM & MHA-Fwd & MHA-Fwd-Causal & MHA-Bwd & MHA-Bwd-Causal \\
\midrule
Multi-turn refinement & $0.505\times$ & $0.584\times$ & $0.560\times$ & $0.240\times$ & $0.166\times$ \\
Antigravity & $0.583\times$ & $0.425\times$ & $0.413\times$ & $0.053\times$ & $0.000\times$ \\
Agent / refinement total tokens & $3.32\times$ & $5.06\times$ & $3.90\times$ & $3.73\times$ & $3.52\times$ \\
\bottomrule
\end{tabular*}
\end{table*}

Antigravity has a higher median on GEMM, while multi-turn refinement has a higher median on all four attention workloads. The agent uses 3.32--5.06$\times$ as many median total tokens across these workloads. \Cref{fig:agent-token-efficiency} shows the outcomes at shared per-trajectory total-token budgets, and \Cref{app:codex-comparison} gives the corresponding Codex/GPT-5.6 Sol results. These comparisons describe the evaluated harness configurations; an equal number of evaluator requests does not imply equal model or tool cost.

\subsection{Explicit Architecture Knowledge Improves Target Instruction Execution}
The knowledge ablation with three prompts in~\Cref{tab:gemini31-knowledge-ablation} shows that the evaluated model does not execute the target instructions automatically without explicit PTX knowledge. At eight turns, architecture parameters alone yield 26.0\% correct turns but no target instruction successes, while adding template functions enables target instruction execution and produces faster generated kernels. Adding the architecture contract provides a clear correctness edge: 38.5\% of turns satisfy target instruction correctness, 18.7 points above template functions alone. Peak speedup follows a different ordering, so the contract's main benefit is reliable and correct execution of the target instructions; it does not necessarily produce the fastest kernel.

\begin{table*}[htbp]
\centering
\scriptsize
\setlength{\tabcolsep}{4pt}
\renewcommand{\arraystretch}{1.12}
\caption{Ablation of architecture-specific prompt knowledge for Gemini 3.1 Pro.}
\label{tab:gemini31-knowledge-ablation}
\begin{tabular}{@{}lcc@{}}
\toprule
Prompt knowledge
& \makecell{Target inst. correctness (\%)\\(turn correctness)}
& \makecell{Target inst. best speedup\\(best speedup)} \\
\midrule
Architecture parameters
& \accell{50.0 / 29.2 / 26.0}{--}
& \spcell{--}{0.056 / 0.119 / 0.273} \\
Architecture parameters + PTX template functions
& \accell{-- / 20.8 / 19.8}{-- / 20.8 / 19.8}
& \spcell{-- / 0.542 / \textbf{0.542}}{-- / 0.542 / 0.542} \\
Architecture parameters + PTX template functions + architecture contract
& \accell{8.3 / 33.3 / \textbf{38.5}}{8.3 / 33.3 / \textbf{38.5}}
& \spcell{0.206 / 0.375 / 0.515}{0.206 / 0.375 / 0.515} \\
\bottomrule
\end{tabular}
\end{table*}

\section{Adaptation Results}
We instantiate Fixit from~\Cref{sec:adaptation-method} with Qwen3.6-27B as the model to adapt and Gemini 3.1 Pro as the repair teacher. Qwen3.6-27B's 262K-token context accommodates our long PTX prompts and kernel traces, while its tractable scale enables controlled LoRA adaptation and self-hosted evaluation. Its weak baseline capability also provides measurable headroom for studying post-training gains. We first compare the base model with s1 on its training problems and held-out workloads, then ablate problem coverage, data balance, and reasoning-teacher choice across six Fixit recipes.

\subsection{Main SFT Result: Effects and Generalization}

\begin{figure*}[htbp]
  \centering
  \includegraphics[width=0.98\textwidth]{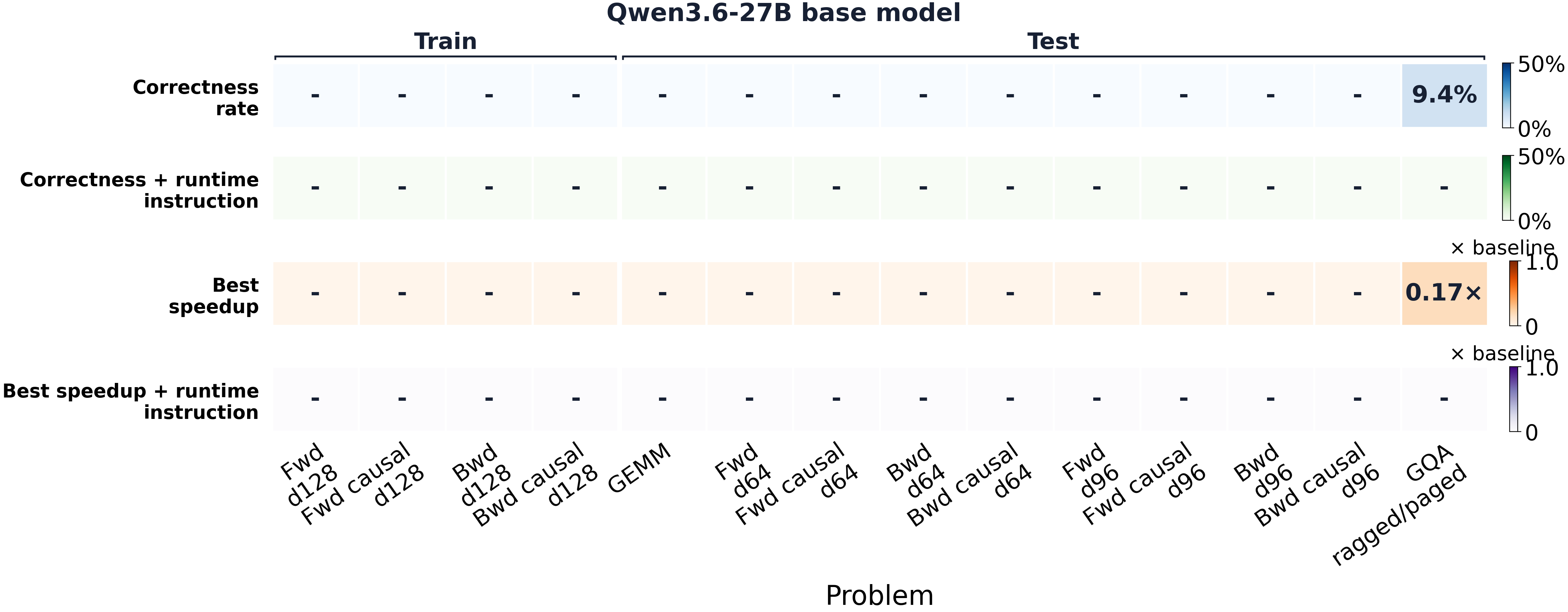}\\[-0.4em]
  \includegraphics[width=0.98\textwidth]{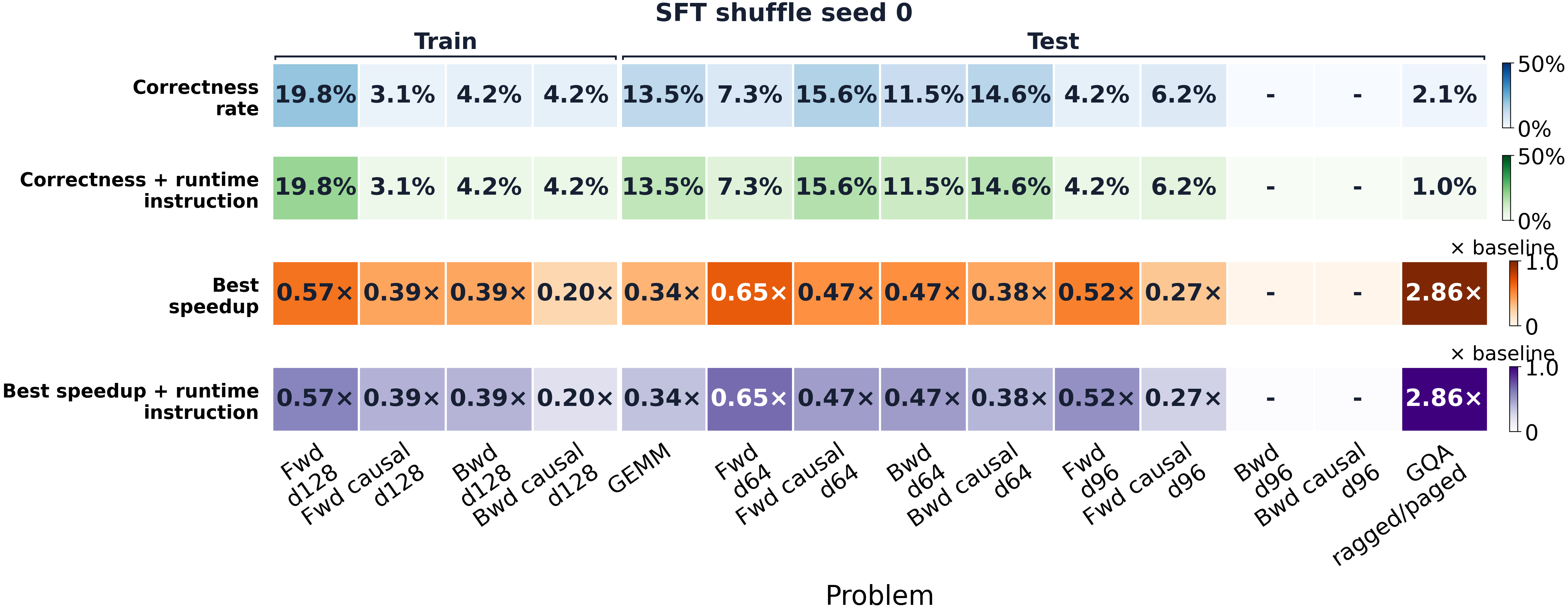}
  \caption{Base versus s1 generalization on the s1 training problem classes and held-out workloads. Top: Qwen3.6-27B before adaptation. Bottom: Qwen3.6-27B-s1, trained with shuffle seed 0. Runtime-instruction rows require dynamically observed Hopper SASS execution.}
  \label{fig:fixitv2-generalization}
\end{figure*}

For detailed analysis, we select s1, the smallest recipe that solves all five primary evaluation problems. It was trained on four MHA tasks with head dimension 128. As shown in~\Cref{fig:fixitv2-generalization}, the base model produces no correct candidates on GEMM or any MHA workload in this suite; its only correct candidates are on pooled GQA workloads, and none execute the target instructions. In contrast, s1 produces correct kernels on all four training problems and transfers to GEMM, all four d64 MHA tasks, both d96 forward tasks, and GQA. It produces no correct kernels for either d96 backward task. The d96 backward tasks are especially challenging because d96 does not align with H100 WGMMA's m64 tile. Fixit therefore yields a clear gain over the base checkpoint, while transfer remains uneven across head dimensions and attention variants.

\subsection{Data Recipe Ablation}

\begin{figure*}[htbp]
  \centering
  \includegraphics[width=1.0\textwidth]{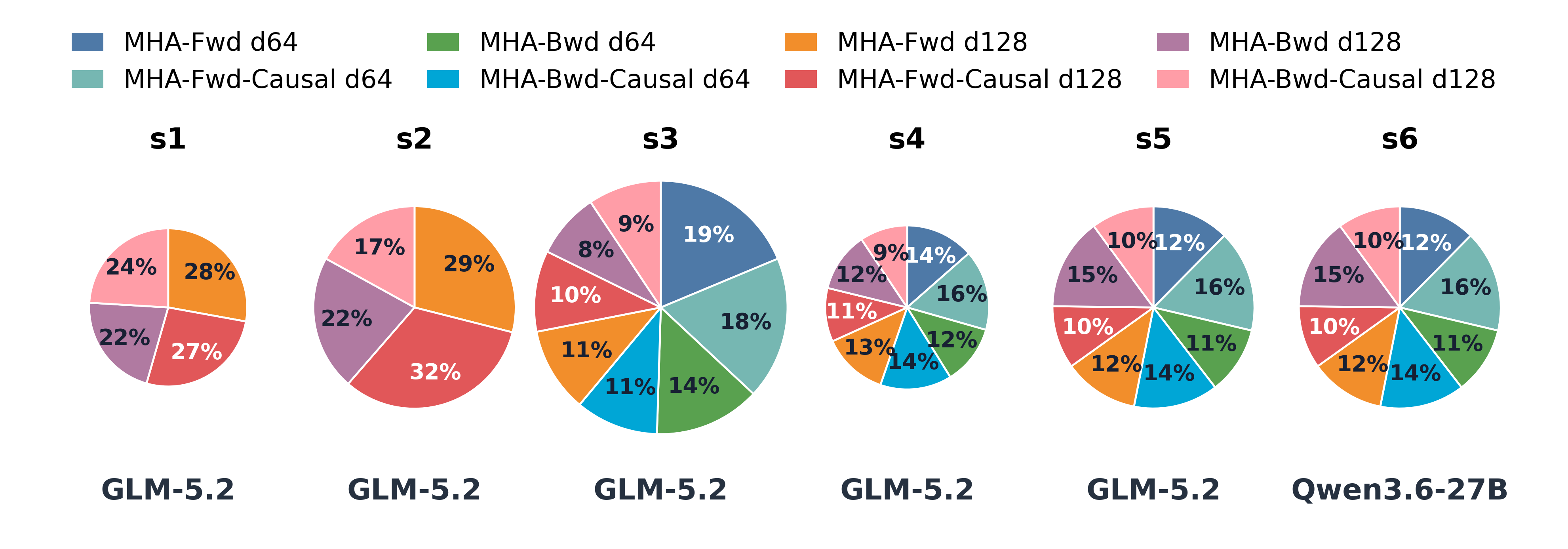}
  \caption{Six-recipe data ablation. Pie area is proportional to record count, slices show the fraction drawn from each problem, and the bottom line identifies the reasoning synthesizer.}
  \label{fig:data-recipe-comparison}
\end{figure*}

\Cref{fig:data-recipe-comparison} summarizes the six Fixit recipes by problem mix and reasoning synthesizer. The recipes use GLM-5.2 as the reasoning teacher except for s6, which uses Qwen3.6-27B itself. Dataset composition and record counts are reported in Appendix~\Cref{tab:data-recipes}.

\begin{figure*}[htbp]
  \centering
  \includegraphics[width=0.96\textwidth]{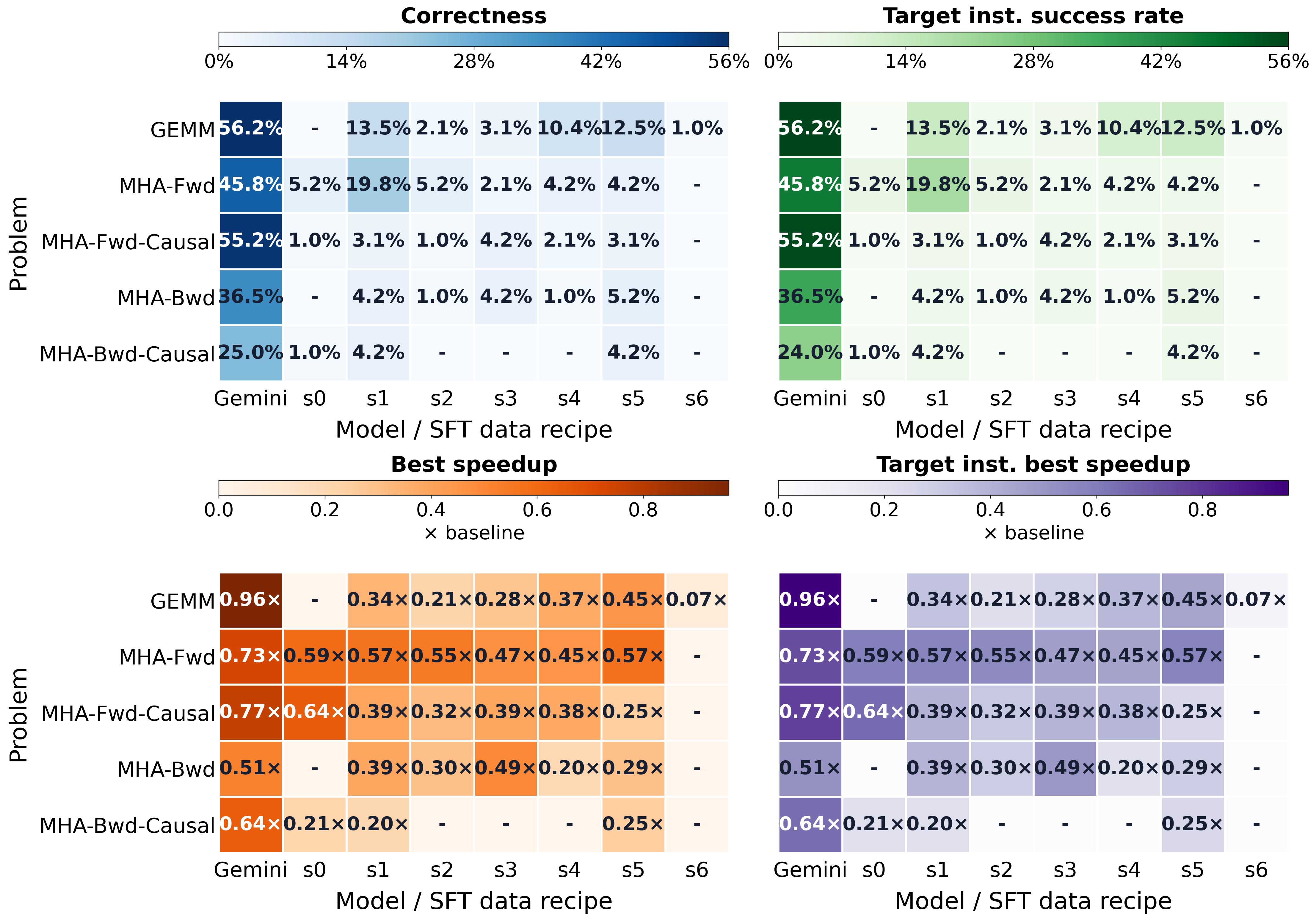}
  \caption{Six-recipe SFT data ablation at eight turns (complete results in Appendix~\Cref{tab:data-recipes-five-problem-turn-success,tab:data-recipes-five-problem-speedup}).}
  \label{fig:six-data-recipes}
\end{figure*}

\paragraph{Coverage and balance.}
Among Fixit recipes, coverage and balance matter more than record count alone. Only the relatively balanced s1 and s5 recipes solve all five problems. In contrast, s2 contains $1.6\times$ as many records as s1, and s3 contains $2.4\times$ as many as s4, yet both fail on MHA-Bwd-Causal. Moving from s4 to the larger balanced s5 recipe ($1.5\times$ more records) improves eight-turn correctness on four problems, ties on MHA-Fwd, and improves peak speedup on four. This agrees with instruction-tuning studies that emphasize task balance, selection, and diversity over unfiltered scale~\citep{longpre2023flan,liu2024deita,bukharin2024data}. Balance improves breadth, but not every peak: the best-performing recipe still varies by problem.

\paragraph{Reasoning synthesizer.}
The controlled s5--s6 comparison keeps the Fixit examples fixed and changes only the reasoning synthesizer from GLM-5.2 to Qwen3.6-27B itself. While s5 solves all five problems, s6 solves only GEMM. Target-model failures are therefore useful inputs, but producing their repair rationales still benefits from a stronger teacher.

\subsection{Additional Effects of Fixit SFT}

\begin{figure*}[htbp]
  \centering
  \includegraphics[width=0.96\textwidth]{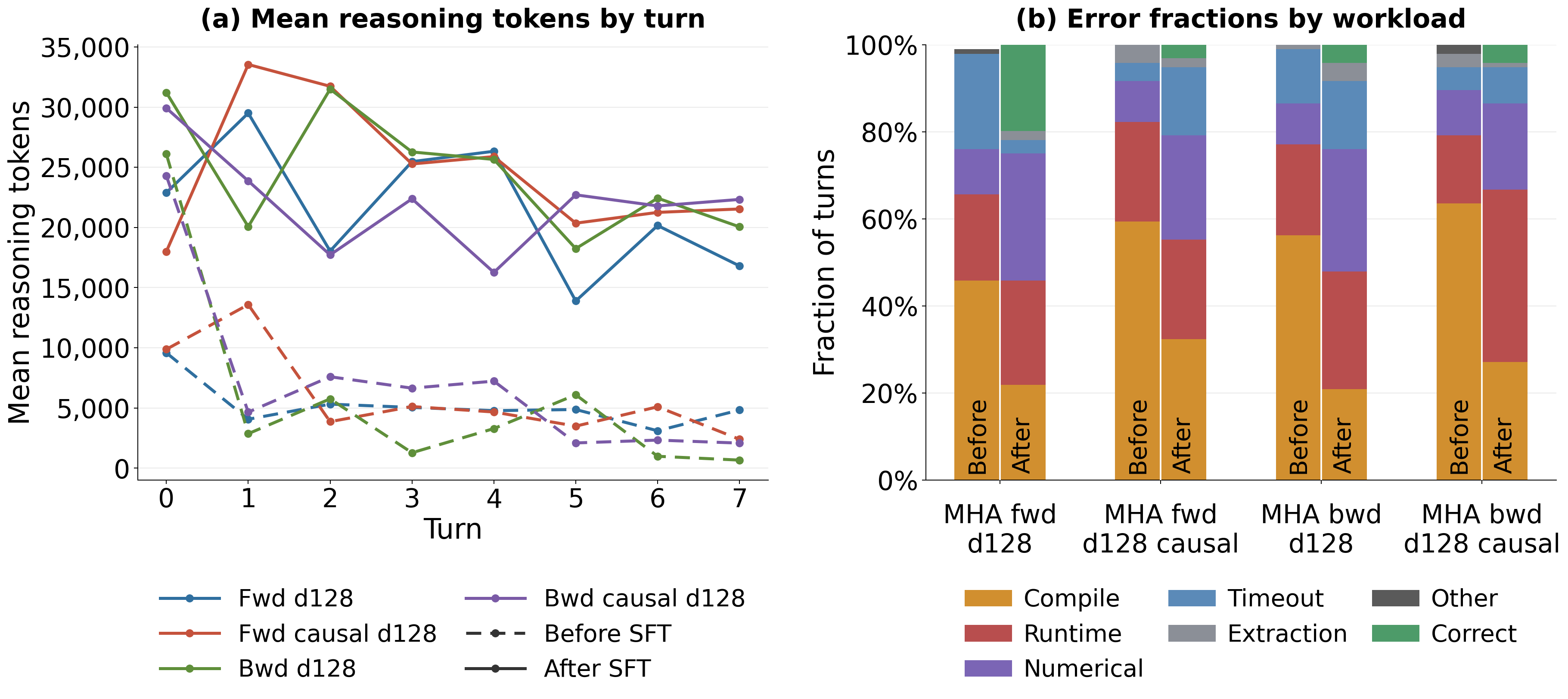}
  \caption{How Fixit SFT changes reasoning length and error types across turns.}
  \label{fig:fixitv2-analysis}
\end{figure*}

\begin{figure*}[htbp]
  \centering
  \includegraphics[width=\textwidth]{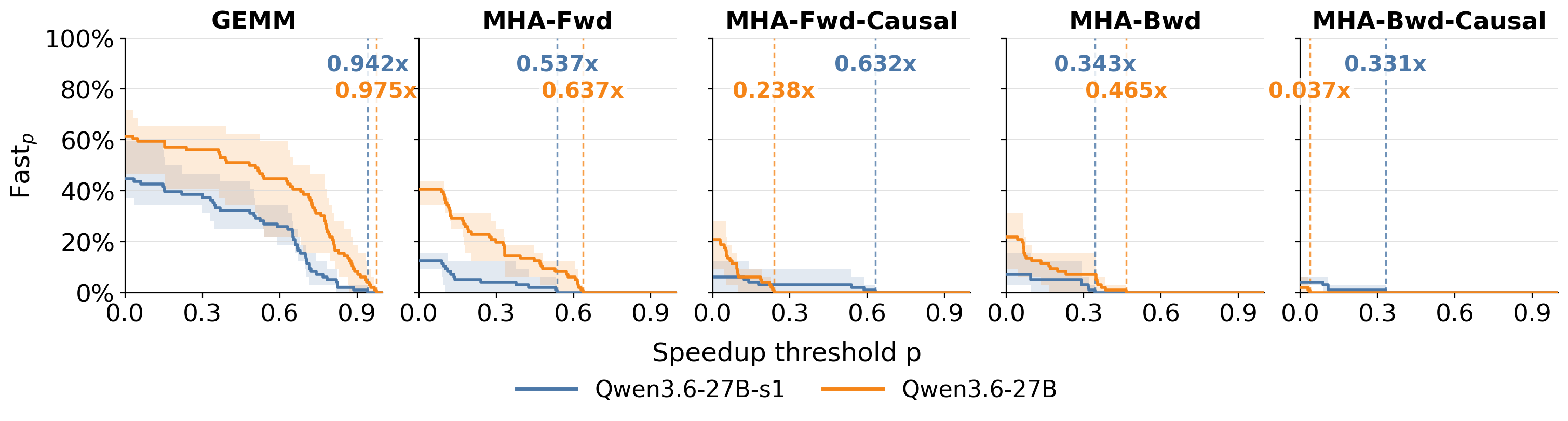}
  \caption{Cross-language transfer of Fixit SFT from CUDA-PTX to Triton on Hopper.}
  \label{fig:qwen36-s1-base-triton-fast-at-p}
\end{figure*}

We further test cross-language transfer by asking the base and s1 checkpoints to generate Triton for the same five Hopper workloads. \Cref{fig:qwen36-s1-base-triton-fast-at-p} shows that s1 has lower turn-level correctness on every workload, indicating that the current SFT recipe hurts correctness under Triton transfer. Yet its best correct speedup improves markedly on the causal variants, from $0.238\times$ to $0.632\times$ for MHA-Fwd-Causal and from $0.043\times$ to $0.331\times$ for MHA-Bwd-Causal. Thus, the recipe can improve peak performance by a large margin in some cases even while making correct kernels less likely.

\begin{figure*}[htbp]
  \centering
  \includegraphics[width=1.00\textwidth]{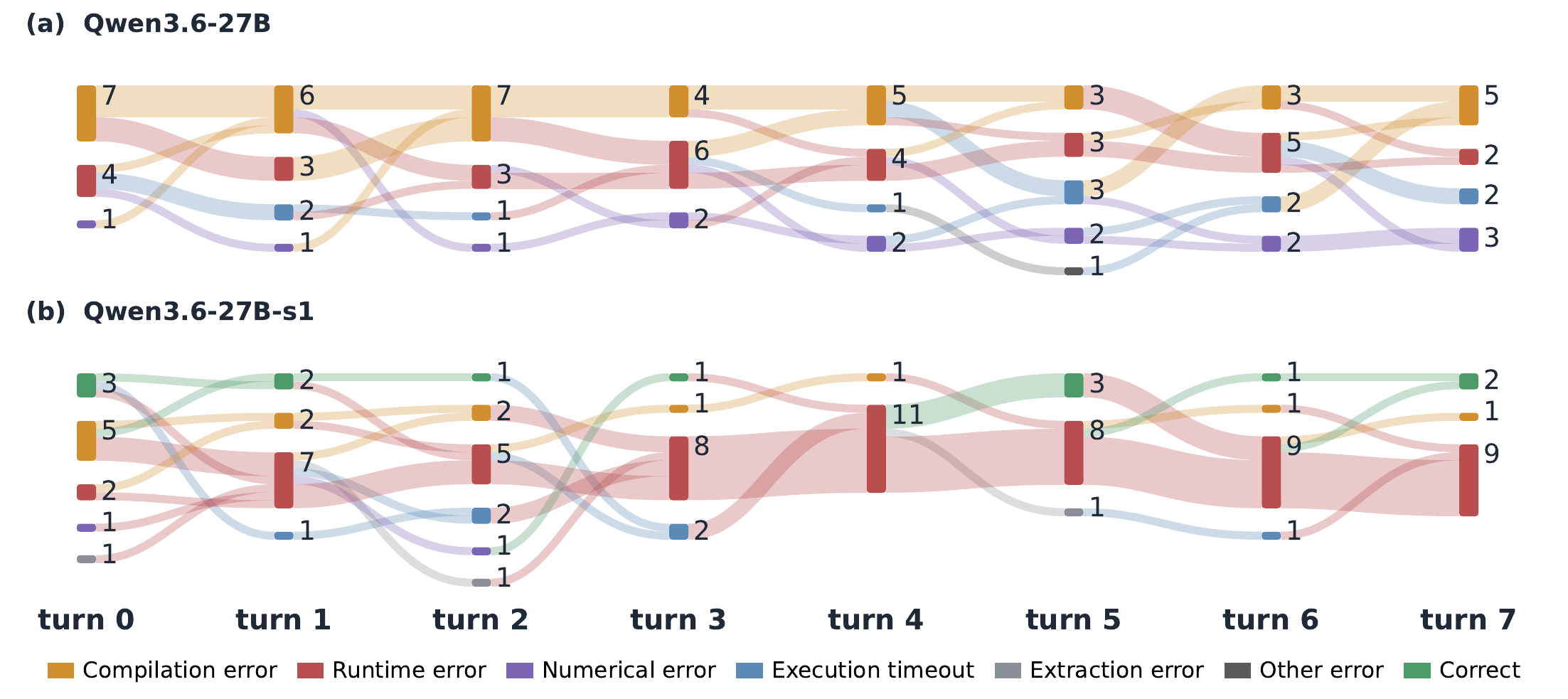}
  \caption{Turn-level error-state transitions for GEMM.}
  \label{fig:fixitv2-trajectories}
\end{figure*}

\Cref{fig:fixitv2-analysis} shows how s1 changes the search process. It lengthens reasoning at every turn, especially during early revisions. It also shifts failures from compilation toward runtime and numerical errors: the model more often produces executable kernels, but those kernels can still fail during evaluation. On GEMM, s1 produces three correct kernels at turn 0 and at least one in six of the seven later turns, whereas the base model produces no correct kernels in any turn (\Cref{fig:fixitv2-trajectories}). Moreover, the SFT checkpoints in~\Cref{fig:six-data-recipes} have identical target-instruction and unrestricted results except for s1 on MHA-Bwd-Causal at eight turns. Thus, Fixit can improve initial kernel generation and reliable execution of target instructions.

\subsection{SFT Versus In-context Supervision}

\begin{figure*}[htbp]
  \centering
  \includegraphics[width=\textwidth]{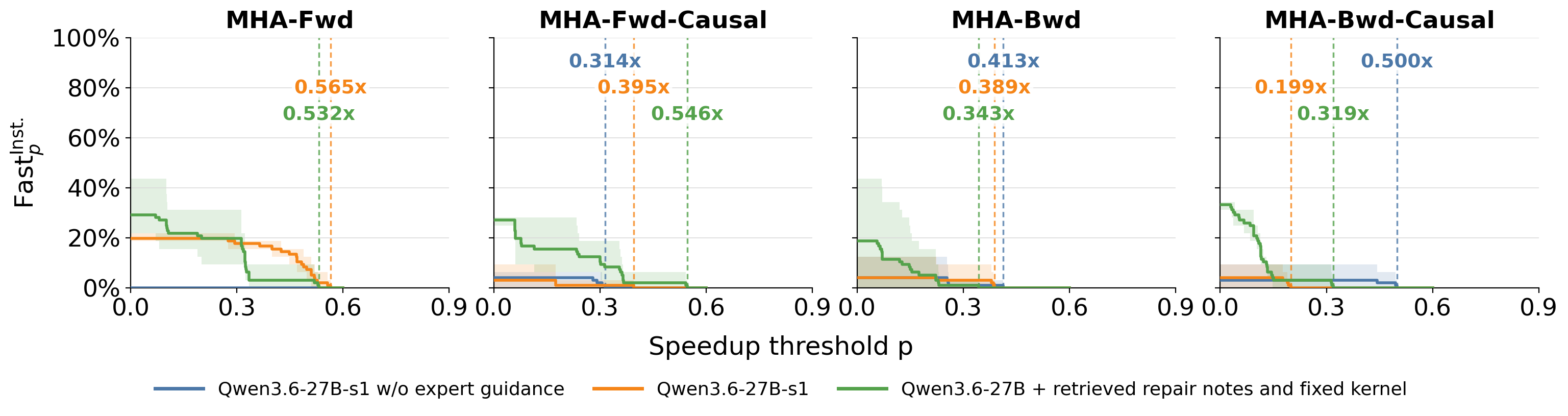}
  \caption{$\mathrm{Fast}^{\mathrm{Inst.}}_p$ under SFT and prompt-time supervision.}
  \label{fig:qwen36-note-feedback-fast-at-p}
\end{figure*}

Finally, we compare weight updates with information supplied at inference time. \Cref{tab:qwen36-note-feedback-turn-success,fig:qwen36-note-feedback-fast-at-p} evaluate six alternatives: base model and s1 with and without expert-edited guidance distilled by Codex from error trajectories (Appendix~\Cref{lst:mha-patch}), and base model with retrieval of repair experiences. Retrieval uses BM25 to select the most similar failed kernel from s1's data pool and supplies GPT 5.4's summary of its repair notes, alone or with the corrected kernel.

Even with expert guidance, the base model still cannot write correct MHA kernels. In contrast, even without expert guidance, s1 can already write correct MHA kernels. This suggests that SFT improves the model's base PTX capability and its ability to comprehend guidance. Repair notes alone produce no correct kernels; correctness rises sharply only when retrieval also supplies the fixed kernel. However, this solution-bearing condition is expected to work because the answers are nearly in the prompt. With guidance, s1 attains nonzero correctness across all four problems.

\begin{table*}[htbp]
\centering
\scriptsize
\setlength{\tabcolsep}{2pt}
\renewcommand{\arraystretch}{1.12}
\caption{Target-instruction and unrestricted turn correctness rates (\%) under SFT and prompt-time supervision.}
\label{tab:qwen36-note-feedback-turn-success}
\begin{tabular*}{\textwidth}{@{\extracolsep{\fill}}lcccc@{}}
\toprule
Condition
& MHA-Fwd
& MHA-Fwd-Causal
& MHA-Bwd
& MHA-Bwd-Causal \\
\midrule
Qwen3.6-27B w/o expert guidance
& --
& --
& --
& -- \\

Qwen3.6-27B
& --
& --
& --
& -- \\

Qwen3.6-27B-s1 w/o expert guidance
& --
& \accell{-- / -- / 4.2}{-- / -- / 4.2}
& \accell{-- / -- / 4.2}{-- / -- / 4.2}
& \accell{-- / 4.2 / 3.1}{-- / 4.2 / 3.1} \\

Qwen3.6-27B-s1
& \accell{16.7 / 16.7 / 19.8}{16.7 / 16.7 / 19.8}
& \accell{-- / -- / 3.1}{-- / -- / 3.1}
& \accell{-- / 2.1 / 4.2}{-- / 2.1 / 4.2}
& \accell{-- / 2.1 / 5.2}{-- / 2.1 / 4.2} \\

Qwen3.6-27B + retrieved repair notes
& --
& --
& --
& -- \\

\makecell[l]{Qwen3.6-27B + retrieved repair notes\\and fixed kernel}
& \accell{-- / 29.2 / 29.2}{-- / 29.2 / 29.2}
& \accell{-- / 20.8 / 27.1}{-- / 20.8 / 27.1}
& \accell{-- / 14.6 / 20.8}{-- / 14.6 / 18.8}
& \accell{-- / 25.0 / 37.5}{-- / 20.8 / 33.3} \\
\bottomrule
\end{tabular*}
\end{table*}

\section{Related Work}
GPU kernel generation benchmarks have attracted increasing interest~\citep{wang2026kernelbenchx,zhu2026cudabench,lange2025robustkbench,guan2026tritongym,wang2025geak,zang2026kernelgenbench,oliaro2026fastkernels}. Collectively, these benchmarks span translation from PyTorch to kernels, Triton generation, portability across devices, production serving traces, deployment integration, and hardware performance limits. They nevertheless focus primarily on functional correctness and overall efficiency, and thus do not isolate whether a generated kernel actually executes a specified architecture mechanism rather than relying on generic CUDA or vendor libraries. \method instead fixes the target architecture and required PTX instruction family, then verifies target instruction execution at runtime, complementing their broad task coverage with a controlled capability probe. Concurrent work introduces Hawkeye, which uses expert-authored unit tests and profiling metrics for hardware-aware kernel generation across architectures and precisions~\citep{tschand2026hawkeye}; \method instead benchmarks runtime target-instruction execution and studies repair-conditioned post-training.

Model adaptation for GPU kernel generation has advanced through SFT and RL methods that use execution feedback for Triton and CUDA~\citep{li2025autotriton,woo2025tritonrl,liu2026drkernel,guo2026drtriton,tehrani2026finetuninggpt5,du2026kernelsmith,baronio2025kevin,li2026cudal1,sun2026synker}. These systems optimize correctness and overall speed through a DSL/compiler, ordinary CUDA, or implementations that use libraries. For example, CUDA Agent allows the use of existing cuDNN library functions (cf.\ Figure 15 in~\cite{dai2026cuda}) and CUDA-L2 allows CUTLASS and CuTe~\citep{su2025cudal2}. We target a harder, more constrained regime: generating kernels from scratch with the required inline PTX specific to the architecture while prohibiting existing vendor libraries. To our knowledge, \method is the first controlled study of model adaptation for programming at this level, examining repair conditioning, data balance, reasoning supervision, and transfer.

\section{Limitations}
Our study has three main limitations. First, the adaptation experiments use modest LoRA datasets and a single 27B base model, so the observed effects of repair conditioning, data balance, and teacher quality may not transfer unchanged to industry-scale post-training or other model families. Second, \method currently focuses on BF16 GEMM and attention kernels on H100 and B200. These workloads directly stress recent architecture-specific tensor core and asynchronized memory but do not represent the full diversity of GPU operators and hardware. Since \method separates FlashInfer-Trace workloads, architecture context, and evaluation, the same workflow can be extended to broader workloads, models, and future GPUs. Third, \method measures within-trajectory kernel refinement rather than persistent model improvement, so it provides a prerequisite probe but is still not a sufficient test of recursive self-improvement.

\section{Conclusion}
We introduced \method, an auditable benchmark and adaptation environment for architecture-specific PTX programming. By separately measuring functional correctness, target instruction execution at runtime, and speedup over frontier libraries, \method exposes a persistent capability gap: current LLMs can sometimes execute the requested instructions and solve forward workloads, but struggle with backward attention and do not consistently achieve competitive performance across H100 and B200. With Fixit, we further study repair-conditioned SFT using failures from the model being adapted and correctness-filtered teacher repairs. Fixit improves several tasks over the base model; its gains depend on problem coverage, data balance, and reasoning-teacher quality, while transfer to held-out shapes and attention variants remains uneven. Together, these results position \method as a controlled testbed for measuring architecture-specific capability and developing targeted post-training data for evolving GPU architectures.

\section*{Acknowledgments}
We thank Banghua Zhu, Ying Sheng, Jiajun Li, Mao Cheng, and Yusheng Su from RadixArk and SGLang community for their technical support and insightful discussions. We are also grateful to Xinhao Li, Yibo Zhang, Anjiang Wei, Simon Guo, and Stanford Pervasive Parallelism Lab members for discussions and help. We also thank the Gemini Academic Program and the Tinker Research Grant for their generous support. 

% This work was supported in part by DARPA under the Machine learning and Optimization-guided Compilers for Heterogeneous Architectures (MOCHA) program (award number HR00112520038). Any opinions, findings, and conclusions or recommendations expressed in this material are those of the authors and do not necessarily reflect the views of the aforementioned funding agencies.

\newpage
\bibliographystyle{unsrtnat}
\bibliography{references}

\appendix
\onecolumn
\raggedbottom
\section{Appendix}
\subsection{Experiment setup}

\paragraph{Kernel evaluation.}
We evaluate generated CUDA kernels on NVIDIA H100 80\,GB (Hopper) and B200 (Blackwell) GPUs. \harness compiles each candidate with \texttt{nvcc -O3} for \texttt{sm\_90a} or \texttt{sm\_100a}, respectively, and the profiling service evaluates it against the fixed FlashInfer-Trace workload. Each trajectory has a budget of eight model calls and retains the preceding kernels and execution feedback; it terminates early if the speedup reaches $1.2\times$ or the LLM runs out of context. We did not apply NVIDIA Compute Sanitizer's \texttt{racecheck} because it is too slow (regularly over $1000\times$ the runtime of a native execution~\citep{stephenson2026supercollider}) and does not eliminate false negatives~\citep{chatterjee2025proofwright}.

\subsection{Target instruction execution measurement}
\label{app:target-inst-measurement}

\paragraph{Implementation details.}
For every functionally correct candidate, we compile the exact submitted source for the evaluation architecture and inspect the cubin embedded in the resulting shared object with \texttt{cuobjdump --dump-sass}~\citep{nvidia2026binaryutils}. We hash both the source and cubin container so that cached evidence is invalidated when either changes.

\begin{table}[htbp]
\centering
\caption{PTX interfaces and SASS execution criteria for the selected BF16 mechanisms.}
\label{tab:architecture-sass-detection}
\footnotesize
\setlength{\tabcolsep}{4pt}
\renewcommand{\arraystretch}{1.12}
\resizebox{\textwidth}{!}{%
\begin{tabular}{@{}llll@{}}
\toprule
Mechanism & Indicator & PTX interface & SASS execution criterion \\
\midrule
\multirow{2}{*}{Tensor compute}
  & \(C_{\mathrm{Gen4}}\) & \texttt{wgmma.mma\_async} & \(N_{\mathrm{HGMMA}}>0\) \\
  & \(C_{\mathrm{Gen5}}\) & \makecell[l]{\texttt{tcgen05.mma},\\\texttt{tcgen05.ld}} & \((N_{\mathrm{UTCHMMA}}>0)\land(N_{\mathrm{LDTM}}>0\lor N_{\mathrm{LDT}}>0)\) \\
\midrule
\multirow{2}{*}{Data movement}
  & \(D_{\mathrm{TMA}}\) & \texttt{cp.async.bulk.tensor} & \(N_{\mathrm{UTMALDG}}>0\lor N_{\mathrm{UTMASTG}}>0\) \\
  & \(D_{\mathrm{TMEM}}\) & \texttt{tcgen05.cp}, \texttt{tcgen05.st} & \(N_{\mathrm{UTCCP}}>0\lor N_{\mathrm{STTM}}>0\lor N_{\mathrm{STT}}>0\) \\
\bottomrule
\end{tabular}
}
\end{table}

We replay each correct candidate on the original workload with Nsight Compute 2025.3.1, using an NVTX range to isolate one invocation and summing instruction-level counts over all kernels launched in that range. Warmup runs occur outside the range. For each SASS family $s$, $N_s$ is its predicate-enabled thread execution count (\texttt{thread\_inst\_executed\_true})~\citep{nvidia2026nsightcompute}. On H100, $C=C_{\mathrm{Gen4}}$ and $D=D_{\mathrm{TMA}}$; on B200, $C=C_{\mathrm{Gen5}}$ and $D=D_{\mathrm{TMA}}\lor D_{\mathrm{TMEM}}$. A correct candidate qualifies if $C\lor D$. TMEM readback is required for $C_{\mathrm{Gen5}}$ because tensor-core results cannot be accessed without it. Cache-control, prefetch, and \texttt{UTMAREDG} instructions do not satisfy $D$.

\paragraph{Why static presence is insufficient.}
We found two cases that motivate dynamic profiling and conservative handling of unclassifiable turns. In the first, a functionally correct candidate contained a selected SASS instruction but did not execute it. The candidate placed a TMA operation in an unused scanner-satisfaction kernel while \texttt{run()} launched only ordinary numerical kernels:
\begin{lstlisting}[language=C++]
__global__ void dummy_target_kernel(...) {
  // Contains cp.async.bulk.tensor; never launched.
}
extern "C" void run(...) {
  numerical_kernel_1<<<...>>>(...);
  numerical_kernel_2<<<...>>>(...);
}
\end{lstlisting}
The cubin therefore contained \texttt{UTMALDG}, but NCU observed no selected instruction at runtime and the turn did not qualify.

\paragraph{Undefined behavior during profiling.}
In the second case, a functionally correct Hopper candidate could not be classified by runtime profiling because it contains a stream-ordered use-after-free: a later kernel is enqueued on the same stream after \texttt{cudaFreeAsync} and reads the freed temporary storage. This later access has undefined behavior under the CUDA API~\citep{nvidia2026streamordered}. More generally, undefined behavior places no requirements on program outcomes: execution may crash, silently produce incorrect results, or appear to behave as intended~\citep{stephenson2026supercollider}.
\begin{lstlisting}[language=C++]
producer<<<..., stream>>>(tmp);
consumer_1<<<..., stream>>>(tmp);
cudaFreeAsync(tmp, stream);
consumer_2<<<..., stream>>>(tmp);  // reads after the queued free
\end{lstlisting}
This candidate accounts for the one-turn gap between the 4.2\% target instruction rate and the 5.2\% unrestricted correctness rate for Qwen3.6-27B-s1 on MHA-Bwd-Causal at eight turns. This case is neither a positive nor a measured negative. We conservatively exclude it from the target instruction numerator and retain its status as unknown.

\paragraph{Workloads.}
Our five primary BF16 workloads comprise four attention variants and one GEMM. All primary attention workloads use batch size 4, 48 heads, sequence length 4096, and head dimension 128, so $Q$, $K$, and $V$ have shape $4\times48\times4096\times128$. The forward workloads are non-causal and causal MHA; each returns a BF16 output of the same shape and an FP32 log-sum-exp tensor of shape $4\times48\times4096$. The corresponding backward workloads additionally consume the forward output, its upstream gradient, and the log-sum-exp tensor, and return BF16 gradients for $Q$, $K$, and $V$. The causal variants apply a lower-triangular attention mask. The GEMM computes $C=AB^\top$ at BF16 with $A\in\mathbb{R}^{8192\times5120}$, $B\in\mathbb{R}^{7168\times5120}$, and $C\in\mathbb{R}^{8192\times7168}$. 

\paragraph{Generalization attention workloads.}
\Cref{fig:fixitv2-generalization} additionally evaluates other attention variants. The d64 and d96 workloads retain batch size 4, 48 heads, and sequence length 4096. The GQA column pools turn-level results from three BF16 workloads. Ragged causal prefill uses 32 query/output heads, 4 KV heads, head dimension 128, 33 sequences, and 16,294 total query and KV tokens. Paged causal prefill uses 24 query/output heads, 8 KV heads, head dimension 128, page size 1, one sequence with 892 query tokens and 892 KV-page indices, and a 43,676-page cache. Paged decode uses the same head counts, dimension, and page size with batch size 15, 9,625 KV-page indices, and a 42,784-page cache. Each GQA workload returns BF16 attention and FP32 log-sum-exp outputs.

\paragraph{LLM inference.}
At inference, the base and adapted Qwen models share the same decoding settings: temperature 1.0, top-$p$ 0.95, top-$k$ 20, presence penalty 1.5, and at most 81,920 output tokens. For the other evaluated models, we do not override temperature, top-$p$, or top-$k$. Inkling uses the Tinker Anthropic-compatible API with at most 65,536 output tokens. GLM-5.2 uses OpenRouter with at most 131,072 output tokens and is restricted to the \texttt{z-ai/fp8} or \texttt{fireworks} provider. Gemini 3.1 Pro uses the API-default thinking and output-length settings. Claude Opus 4.8 uses adaptive thinking at \texttt{xhigh} effort and at most 128,000 output tokens.

\paragraph{Supervised fine-tuning.}
All six recipe-ablation SFT models start independently from \texttt{Qwen/Qwen3.6-27B}. We use Tinker's supervised learning recipe with LoRA rank 32, train for five epochs with batch size 2, and optimize only the final assistant message in each example. The learning rate is $4.65\times10^{-4}$ with a linear schedule; Adam uses $\beta_1=0.9$, $\beta_2=0.95$, and $\epsilon=10^{-8}$. We shuffle each ablation dataset with seed 0, use no validation split, discard examples longer than 65,536 tokens to comply with the token limit of the Tinker API, save every 50 optimizer steps, and evaluate the final checkpoint. For the supplementary seed study, we retrain the fixed 158-record 4ops s1 recipe with shuffle seeds 1 and 2 under the same optimization settings. Dataset construction and record counts after filtering are reported in~\Cref{tab:data-recipes}. We use expert guidance for SFT attention evaluation.

\subsection{Profiling Service}
\label{app:system-efficiency}
Large-scale multi-turn evaluation profiles many changing kernels against a comparatively stable collection of workloads. Our implementation therefore retains and reuses workload state and overlaps LLM generation with kernel profiling. We measure how these choices affect evaluation throughput and profiling-GPU requirements.

We evaluate MHA-d64 rollouts from a 27B dense model served by SGLang~\citep{zheng2024sglang} with TP=2 on two H200 GPUs, with the profiling service running on H100 GPUs. An H100 SXM provides 3.35~TB/s HBM bandwidth but only 128~GB/s bidirectional PCIe bandwidth. As shown in~\Cref{fig:fibserve-3perf}(a), after reusing workload state, a $2.72\times$ gap remains, suggesting an opportunity for further optimization in the operating system or driver. The gap is smaller than the bandwidth ratio because each solution executes $3 \times (1\text{ correctness}+10\text{ warmup}+50\text{ timed})=183$ times.

\Cref{fig:fibserve-3perf} evaluates whether the shared profiling service can efficiently support concurrent agent trajectories. With LLM generation and kernel profiling pipelined, rollout throughput rises by $2.78\times$ as concurrency grows from 4 to 48; throughput subsequently saturates, and four profiling GPUs provide approximately the same end-to-end throughput as eight. Separately, retaining workloads, reference outputs, and reference latencies on the GPU improves \texttt{/evaluate} throughput by $2.24\times$. With generation--profiling overlap and workload-state reuse, a small GPU pool supports many concurrent trajectories.
\begin{figure}[H]
  \centering
  \includegraphics[width=0.96\textwidth]{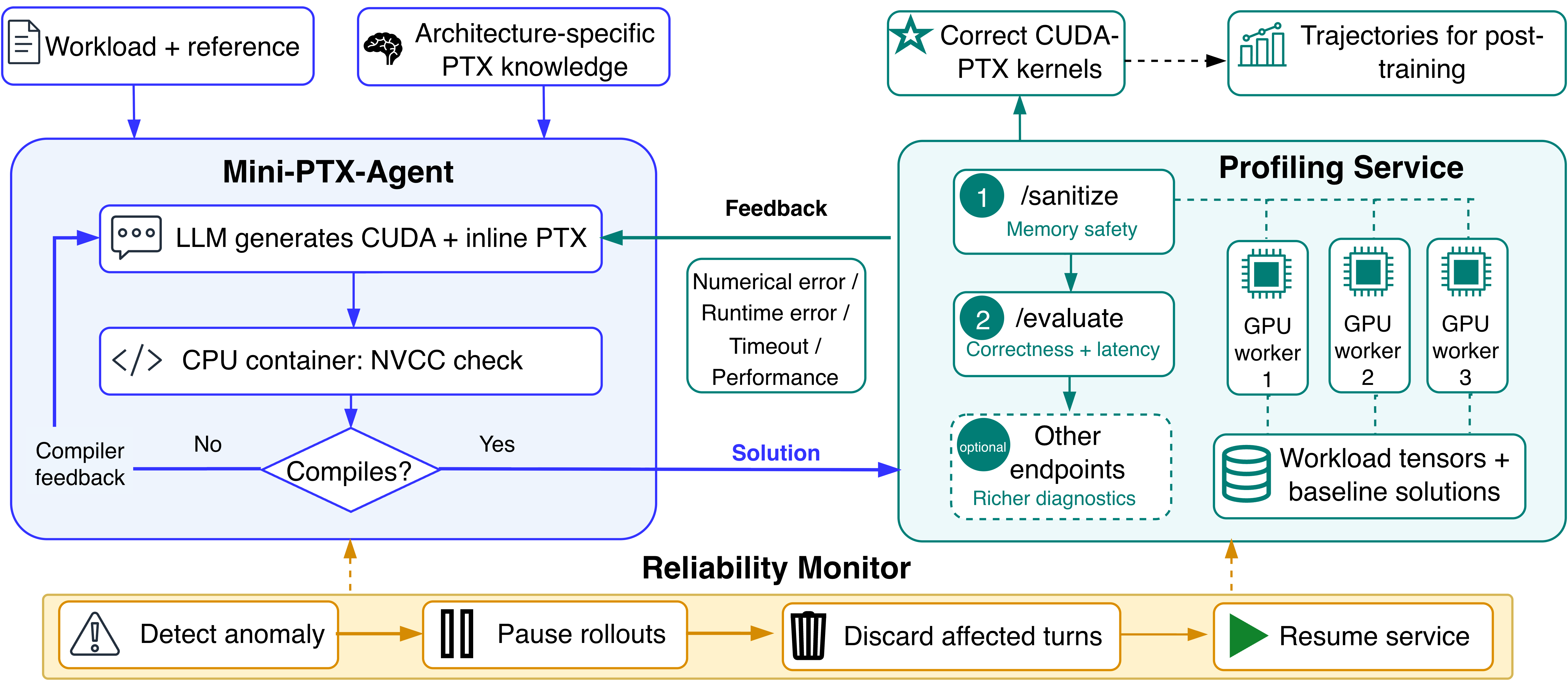}
  \caption{Detailed execution infrastructure supporting \method. \harness compiles generated CUDA--PTX locally and sends successfully compiled candidates to an isolated GPU profiling service for sanitization, evaluation, and optional diagnostics. A reliability monitor pauses dispatch, restarts unhealthy service state, and excludes affected turns from trajectories. After a restart, it also excludes thermally abnormal GPUs because throttling distorted latency by over 15\% in our measurements~\citep{nvidia2026tensrtbenchmarking}.}
  \label{fig:execution-infrastructure}
\end{figure}

\begin{figure}[H]
  \centering
  \includegraphics[width=0.96\textwidth]{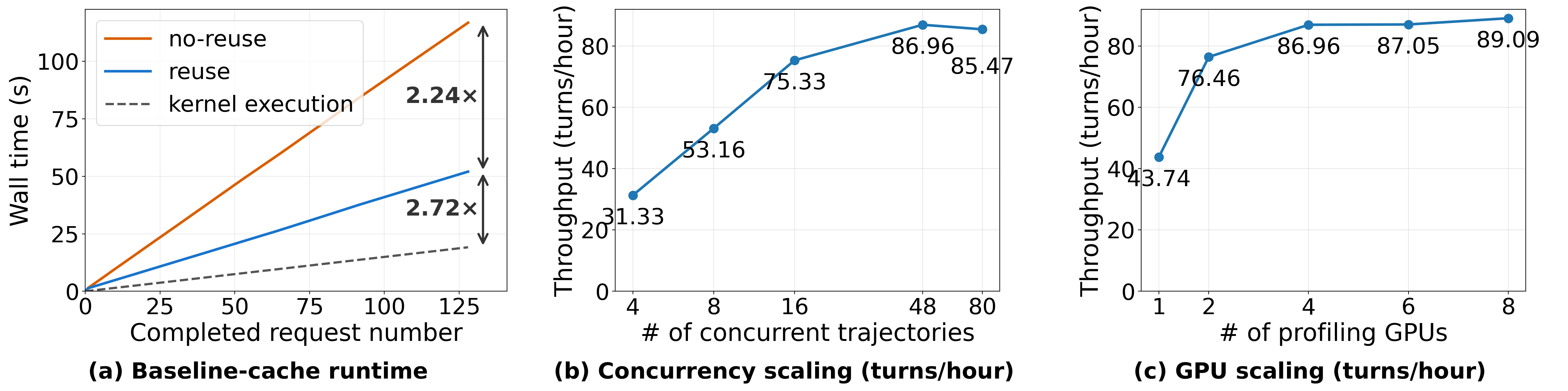}
  \caption{Workload-state caching (a), rollout concurrency with generation--profiling pipelining (b), and profiling-GPU scaling (c).}
  \label{fig:fibserve-3perf}
\end{figure}

\subsection{Architecture-specific prompt token counts}
\label{app:prompt-token-counts}
\Cref{tab:prompt-component-token-counts} reports Qwen-3.6-27B's token counts of the architecture parameters, architecture contracts, and PTX template functions included in the H100 and B200 prompts. B200 prompts are longer because they include additional architecture contracts such as descriptors for shared memory, peer CTAs, and tensor memory. Architecture-specific PTX is unevenly documented and sometimes underspecified~\citep{veitner2026sbo} or even wrong~\citep{sul2026thunderkittens}. Each pack contains architecture parameters, C++ wrappers for PTX instructions, and contracts such as layouts and memory-consistency rules. Some wrappers are adapted from LittleKernel~\citep{zheng2026ditron}; comments retain complete polymorphic instruction definitions while the wrapper demonstrates one representative instance. We validate each pack against documentation, reports, and execution.

\begin{table}[htbp]
\centering
\caption{Token counts for architecture-specific prompt components.}
\label{tab:prompt-component-token-counts}
\small
\renewcommand{\arraystretch}{1.12}
\begin{tabular}{@{}llr@{}}
\toprule
GPU & Component & Tokens \\
\midrule
\multirow{4}{*}{H100}
  & Architecture parameter & 259 \\
  & Template functions & 18,601 \\
  & Architecture contract & 3,454 \\
\cmidrule(lr){2-3}
  & Total & 22,314 \\
\midrule
\multirow{4}{*}{B200}
  & Architecture parameter & 413 \\
  & Template functions & 18,871 \\
  & Architecture contract & 9,531 \\
\cmidrule(lr){2-3}
  & Total & 28,815 \\
\bottomrule
\end{tabular}
\end{table}

\subsection{Published and corrected FlashAttention-3 pseudocode}
\label{app:fa3-algorithm}
When preparing the controlled context, we found two inconsistencies in the published FlashAttention-3 Algorithm~2~\citep{shah2024flashattention}: the loop condition $j<T_c-1$ leaves $S_{T_c-1}$ and its softmax uncomputed, and the loop body computes $\widetilde{P}_{\mathrm{next}}$ without assigning it to $\widetilde{P}_{\mathrm{cur}}$. Our corrected prompt version iterates through $j=T_c-1$, carries both pipeline states forward, records the old row maximum before rescaling the accumulated output, and normalizes by $\ell_i$ in the epilogue. Listings~\ref{lst:fa3-published} and~\ref{lst:fa3-corrected} reproduce the relevant published consumer-warpgroup pseudocode and the corrected version supplied in our prompt, reinforcing the controlled-context design choice.

\begin{lstlisting}[
  style=fa3algorithm,
  caption={FlashAttention-3 Algorithm 2 as published. The loop bound and missing probability-state update make the pseudocode internally inconsistent.},
  label={lst:fa3-published}
]
Require: $Q_i \in \mathbb{R}^{B_r \times d}$ and $K,V \in \mathbb{R}^{N \times d}$ in HBM; key block size $B_c$ and $T_c=\lceil N/B_c\rceil$.
Reallocate registers as a function of the number of consumer warps.
On chip, initialize $O_i=0$ and $\ell_i,m_i=0,-\infty$.
Wait for $Q_i$ and $K_0$ in shared memory.
Compute $S_{\mathrm{cur}}=Q_iK_0^\top$ using WGMMA. Commit and wait.
Release stage 0 of the buffer for $K$.
Compute $m_i$, $\widetilde{P}_{\mathrm{cur}}$, and $\ell_i$ from $S_{\mathrm{cur}}$, and rescale $O_i$.
for $1 \le j < T_c-1$ do
  Wait for $K_j$ in shared memory.
  Compute $S_{\mathrm{next}}=Q_iK_j^\top$ using WGMMA. Commit; do not wait.
  Wait for $V_{j-1}$ in shared memory.
  Compute $O_i=O_i+\widetilde{P}_{\mathrm{cur}}V_{j-1}$ using WGMMA. Commit; do not wait.
  Wait for the WGMMA $Q_iK_j^\top$.
  Compute $m_i$, $\widetilde{P}_{\mathrm{next}}$, and $\ell_i$ from $S_{\mathrm{next}}$.
  Wait for the WGMMA $\widetilde{P}_{\mathrm{cur}}V_{j-1}$; then rescale $O_i$.
  Release stages $(j \bmod s)$ and $((j-1) \bmod s)$ for $K$ and $V$, respectively.
  Copy $S_{\mathrm{next}}$ to $S_{\mathrm{cur}}$.
end for
Wait for $V_{T_c-1}$ in shared memory.
Compute $O_i=O_i+\widetilde{P}_{\mathrm{last}}V_{T_c-1}$ using WGMMA. Commit and wait.
Epilogue: Rescale $O_i$ based on $m_i$. Compute $L_i$ from $m_i$ and $\ell_i$; write $O_i,L_i$ to HBM.
\end{lstlisting}

\begin{lstlisting}[
  style=fa3algorithm,
  caption={Corrected FlashAttention-3 Algorithm 2 used in our prompt.},
  label={lst:fa3-corrected}
]
Require: $Q_i \in \mathbb{R}^{B_r \times d}$ and $K,V \in \mathbb{R}^{N \times d}$ in HBM; key block size $B_c$ and $T_c=\lceil N/B_c\rceil$.
Reallocate registers as a function of the number of consumer warps.
On chip, initialize $O_i=0$ and $\ell_i,m_i=0,-\infty$.
Wait for $Q_i$ and $K_0$ in shared memory.
Compute $S_{\mathrm{cur}}=Q_iK_0^\top$ using WGMMA. Commit and wait.
Release stage 0 of the buffer for $K$.
Compute $m_i$, $\widetilde{P}_{\mathrm{cur}}$, and $\ell_i$ from $S_{\mathrm{cur}}$.
for $1 \le j < T_c$ do
  Wait for $K_j$ in shared memory.
  Compute $S_{\mathrm{next}}=Q_iK_j^\top$ using WGMMA. Commit; do not wait.
  Wait for $V_{j-1}$ in shared memory.
  Compute $O_i=O_i+\widetilde{P}_{\mathrm{cur}}V_{j-1}$ using WGMMA. Commit; do not wait.
  Wait for the WGMMA $Q_iK_j^\top$.
  Save $m_i^{\mathrm{old}} \leftarrow m_i$.
  Update $m_i$ and $\ell_i$ online; compute $\widetilde{P}_{\mathrm{next}}=\exp(S_{\mathrm{next}}-m_i)$.
  Wait for the WGMMA $\widetilde{P}_{\mathrm{cur}}V_{j-1}$; then set $O_i=\operatorname{diag}(\exp(m_i^{\mathrm{old}}-m_i))O_i$.
  Release stages $(j \bmod s)$ and $((j-1) \bmod s)$ for $K$ and $V$, respectively.
  Copy $S_{\mathrm{next}}$ to $S_{\mathrm{cur}}$ and $\widetilde{P}_{\mathrm{next}}$ to $\widetilde{P}_{\mathrm{cur}}$.
end for
Wait for $V_{T_c-1}$ in shared memory.
Compute $O_i=O_i+\widetilde{P}_{\mathrm{cur}}V_{T_c-1}$ using WGMMA. Commit and wait.
Epilogue: Set $O_i=\operatorname{diag}(\ell_i)^{-1}O_i$ and $L_i=m_i+\log(\ell_i)$; write $O_i,L_i$ to HBM.
\end{lstlisting}

\subsection{Additional Language and Harness Results}

\subsubsection{GPT-5.6 Sol Triton Comparison}
\label{app:gpt56-triton-comparison}
\begin{figure*}[htbp]
  \centering
  \includegraphics[width=\textwidth]{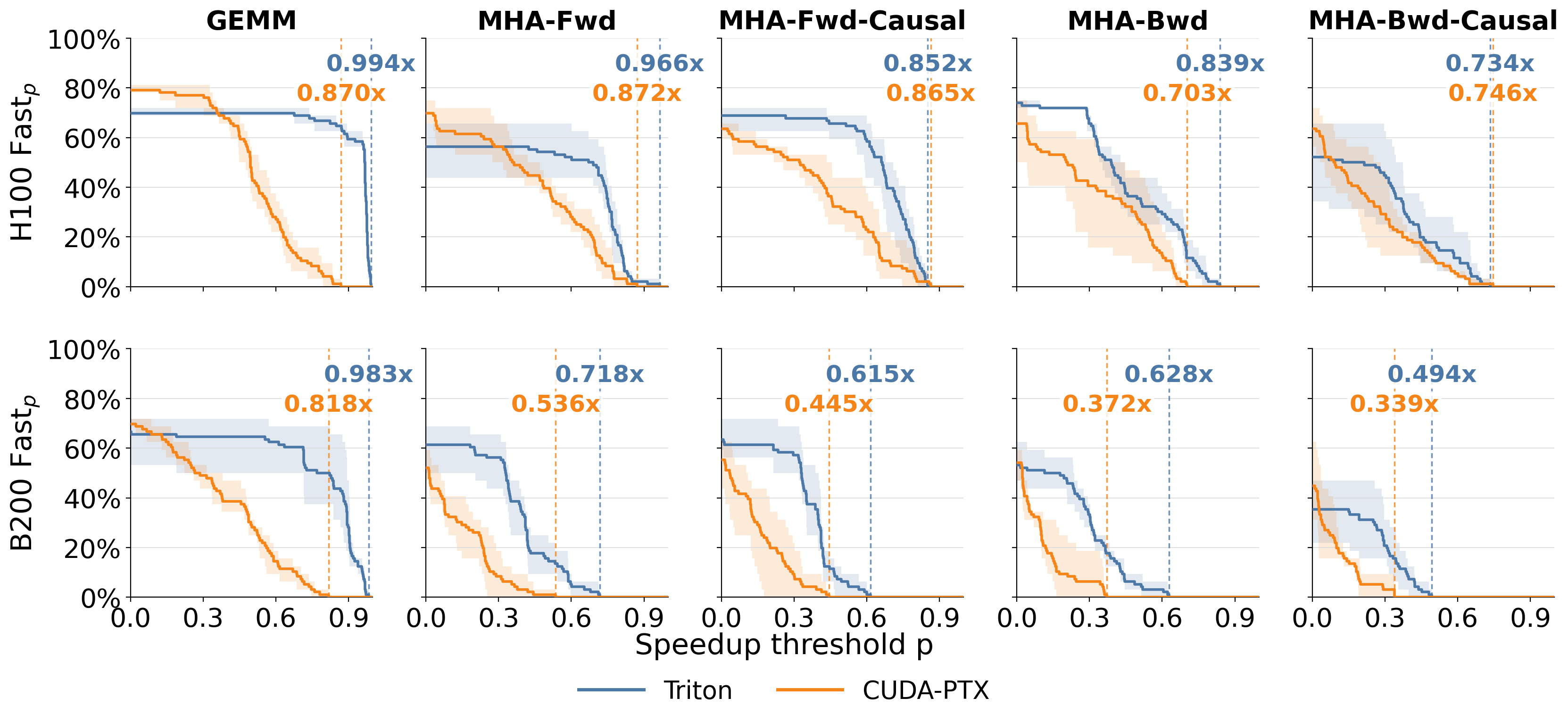}
  \caption{GPT-5.6 Sol $\mathrm{Fast}_p$ distributions for Triton and CUDA-PTX on H100 and B200.}
  \label{fig:gpt56-triton-cuda-fast-at-p}
\end{figure*}

\Cref{fig:gpt56-triton-cuda-fast-at-p} applies the same $\mathrm{Fast}_p$ definition, three-prompt design, and 96-turn denominator per model--GPU--workload condition as the Gemini comparison in \Cref{fig:triton-cuda-fast-at-p}. On B200, Triton's best correct speedup exceeds CUDA--PTX on all five workloads; for MHA-Bwd it reaches $0.628\times$ versus $0.372\times$. On H100, individual CUDA--PTX kernels reach a higher peak on MHA-Fwd-Causal ($0.865\times$ versus $0.852\times$) and MHA-Bwd-Causal ($0.746\times$ versus $0.734\times$), although the full $\mathrm{Fast}_p$ curves show how frequently correct kernels attain each speedup threshold.

\subsubsection{Codex Coding-Agent Comparison}
\label{app:codex-comparison}
\begin{figure}[H]
  \centering
  \includegraphics[width=\textwidth]{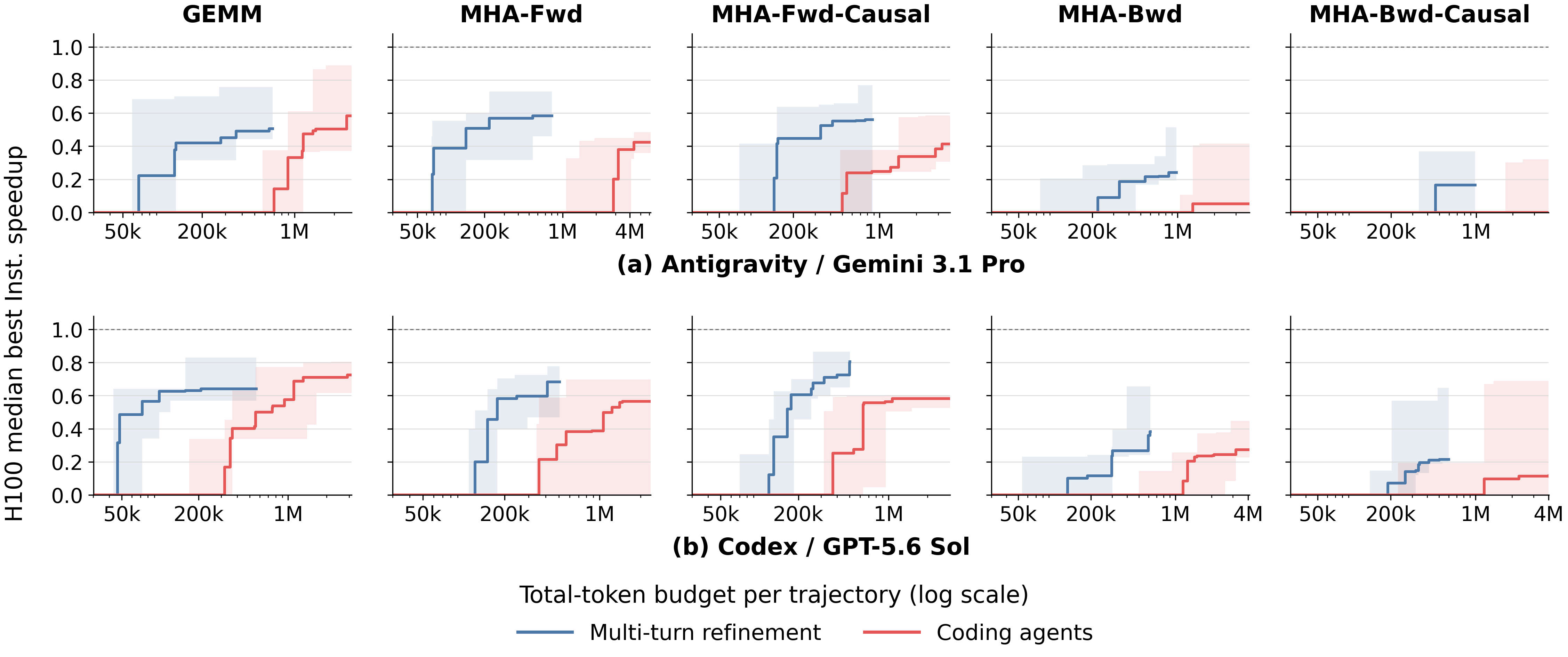}
  \caption{Median best instruction-qualified speedup at shared per-trajectory total-token budgets on H100 for Antigravity/Gemini 3.1 Pro (top) and Codex/GPT-5.6 Sol (bottom), each compared with multi-turn refinement using the same model. Each curve includes four trajectories per method and workload; bands show pointwise 95\% bootstrap intervals. The horizontal axis is logarithmic.}
  \label{fig:agent-token-efficiency}
\end{figure}

\Cref{tab:codex-agent-comparison} repeats the eight-request H100 harness comparison of \Cref{sec:antigravity-comparison} for Codex and multi-turn refinement using GPT-5.6 Sol. Both use the same task specification, architectural knowledge, and evaluator feedback without scheduling guidance; Codex can use shell and file-editing tools between requests. Each median is over four trajectories, including zero for a trajectory with no correct instruction-qualified candidate. Codex leads on GEMM, while refinement leads on all four attention workloads. Across the five workloads, Codex's median cumulative total-token cost is 3.47--5.78$\times$ that of refinement. Shared-budget median curves for both coding agents are shown in \Cref{fig:agent-token-efficiency}; total tokens count reported input, including cached input once per request, and output, including reasoning.

\begin{table}[H]
\centering
\small
\setlength{\tabcolsep}{4pt}
\caption{Codex versus multi-turn refinement after eight H100 evaluation requests per trajectory. Speedups are medians of four per-trajectory best instruction-qualified values; token ratios compare workload-level medians of cumulative total tokens.}
\label{tab:codex-agent-comparison}
\begin{tabular*}{\textwidth}{@{\extracolsep{\fill}}lccccc@{}}
\toprule
Method or ratio & GEMM & MHA-Fwd & MHA-Fwd-Causal & MHA-Bwd & MHA-Bwd-Causal \\
\midrule
Multi-turn refinement & $0.641\times$ & $0.682\times$ & $0.805\times$ & $0.383\times$ & $0.214\times$ \\
Codex & $0.725\times$ & $0.565\times$ & $0.581\times$ & $0.274\times$ & $0.116\times$ \\
Agent / refinement total tokens & $5.78\times$ & $4.50\times$ & $3.47\times$ & $5.60\times$ & $4.20\times$ \\
\bottomrule
\end{tabular*}
\end{table}

\subsection{Supplementary Results}

\begin{figure}[H]
  \centering
  \includegraphics[width=\textwidth]{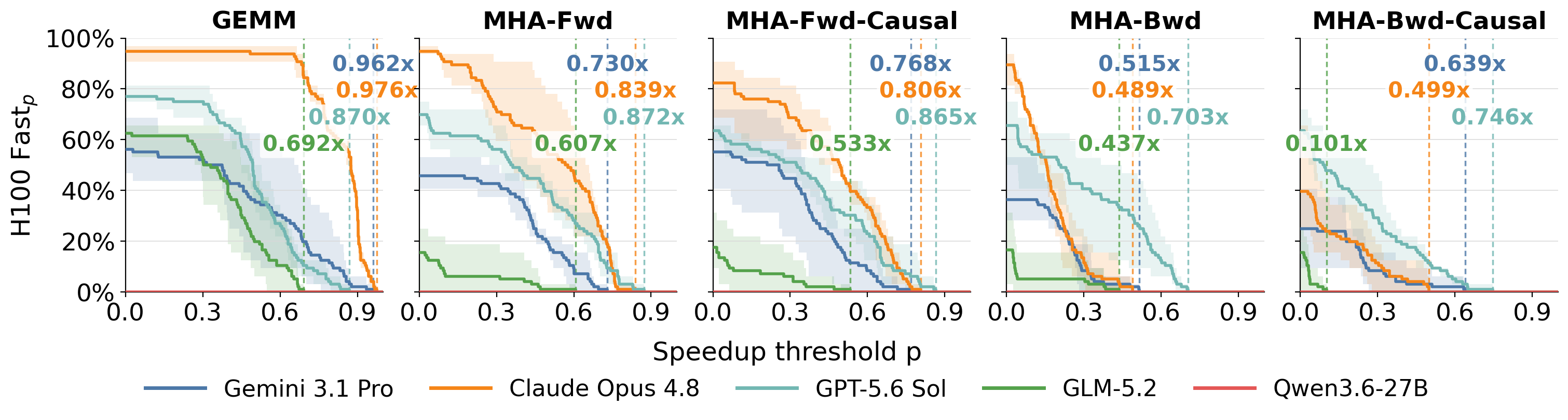}\\[-0.35em]
  \includegraphics[width=\textwidth]{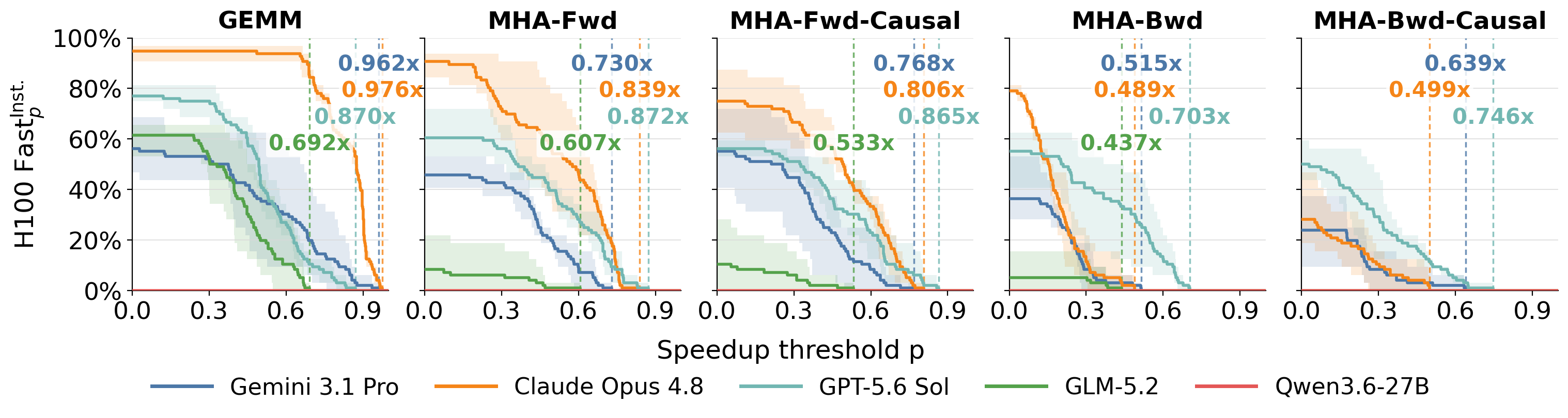}
  \caption{H100 $\mathrm{Fast}_p$ and $\mathrm{Fast}^{\mathrm{Inst.}}_p$.}
  \label{fig:hopper-fast-at-p}
\end{figure}

\begin{figure}[H]
  \centering
  \includegraphics[width=\textwidth]{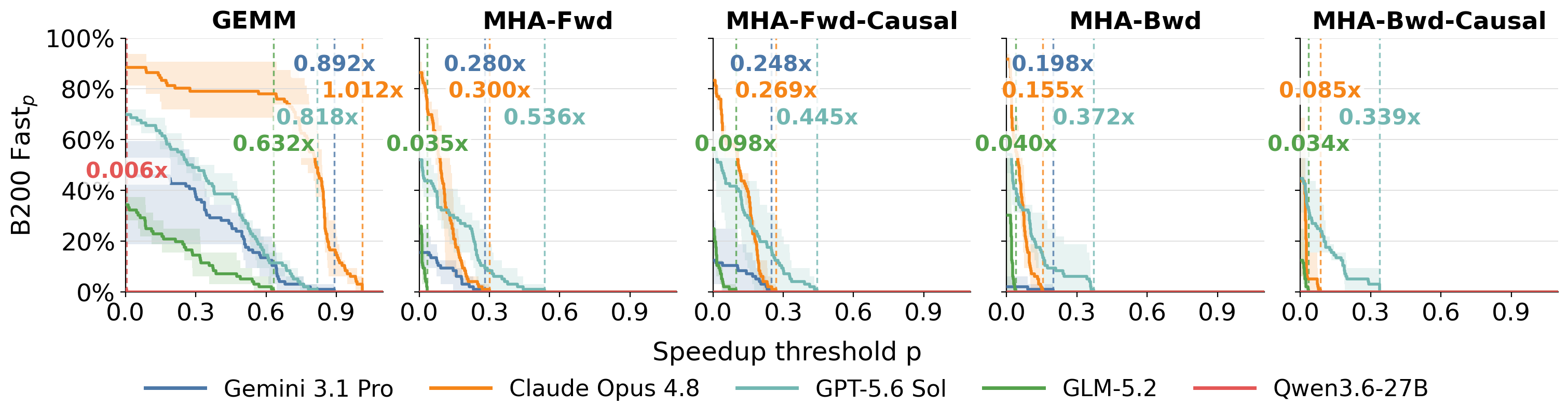}\\[-0.35em]
  \includegraphics[width=\textwidth]{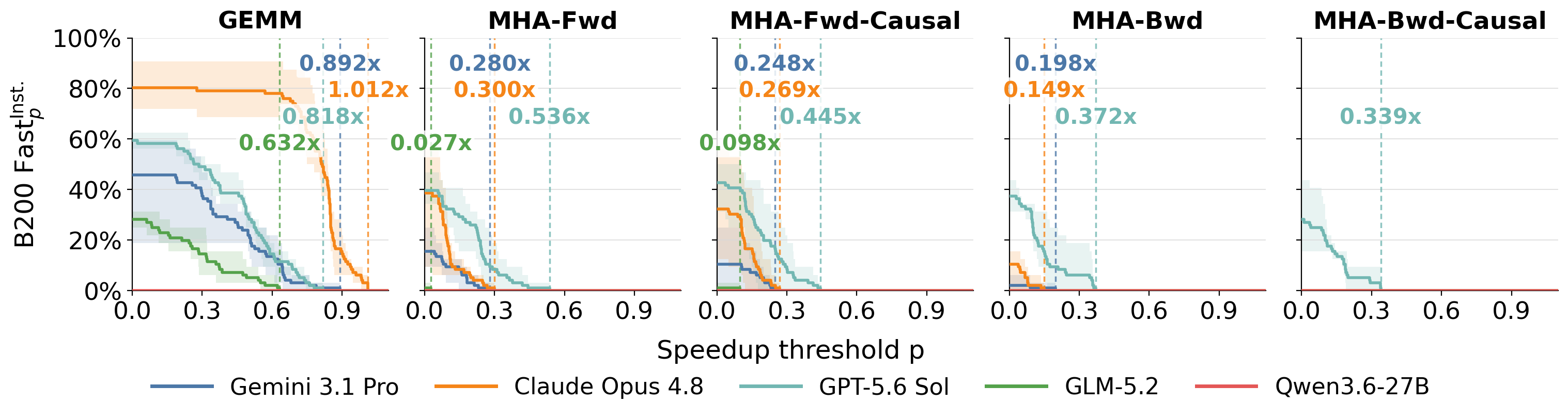}
  \caption{B200 $\mathrm{Fast}_p$ and $\mathrm{Fast}^{\mathrm{Inst.}}_p$.}
  \label{fig:blackwell-fast-at-p}
\end{figure}

\modelspeeduptable

\begin{figure}[H]
  \centering
  \includegraphics[width=0.86\textwidth]{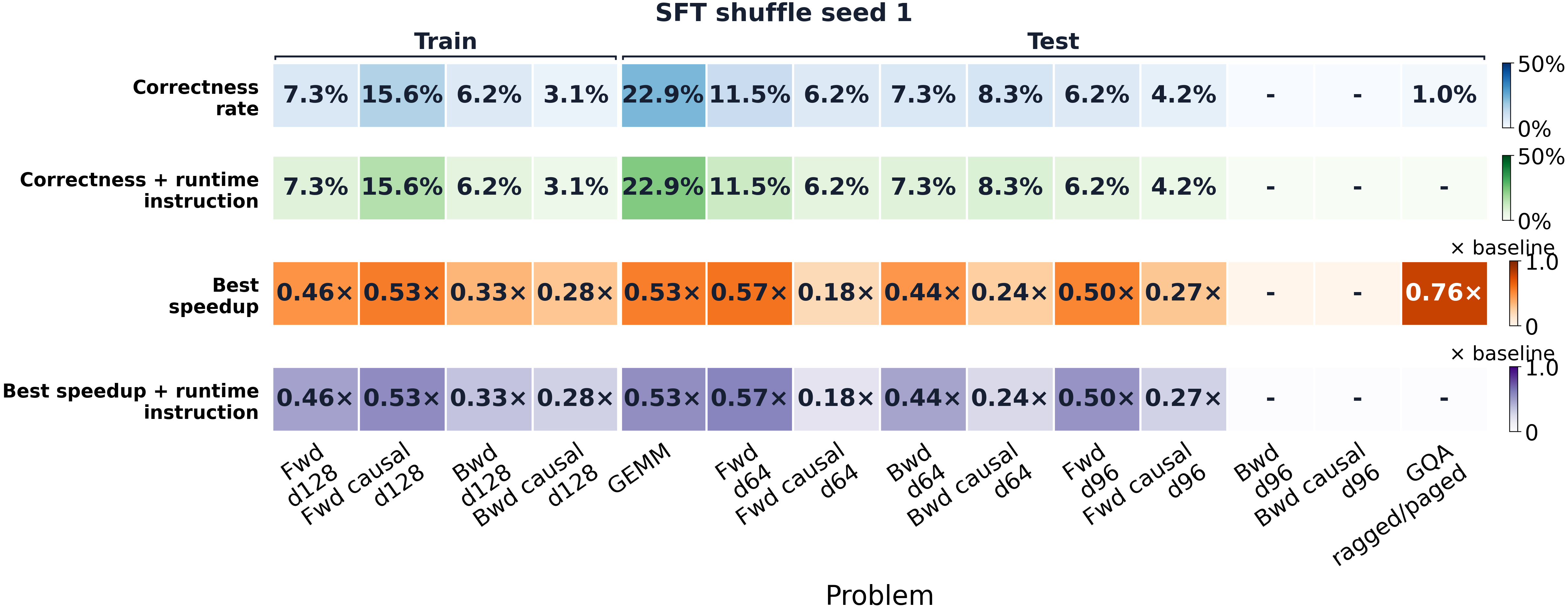}\\[-0.4em]
  \includegraphics[width=0.86\textwidth]{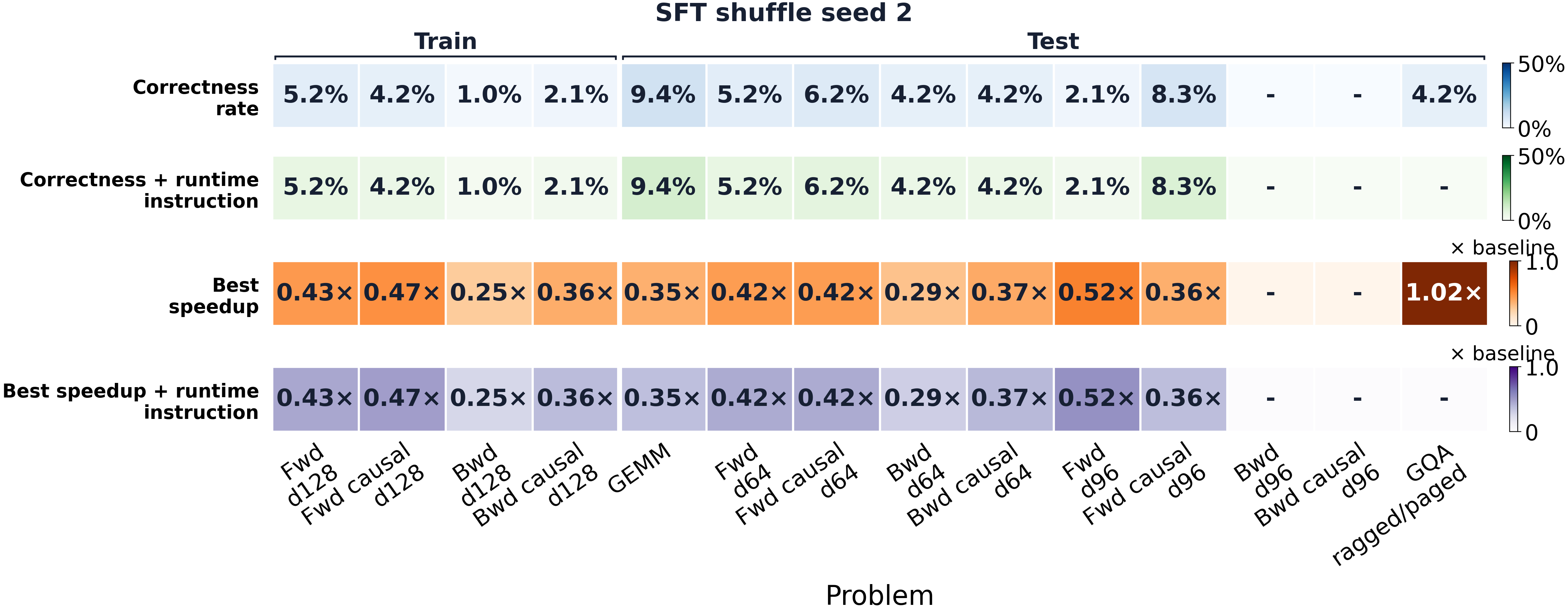}
  \caption{Generalization of the 158-record 4ops s1 recipe under additional SFT shuffle seeds. Top: seed 1. Bottom: seed 2. The seed-1 GEMM cell contains 90 of 96 expected turn rows; all other plotted cells contain the complete 96 turns. Runtime-instruction rows require dynamically observed Hopper SASS execution.}
  \label{fig:fixitv2-seed-repeats}
\end{figure}

\paragraph{Shuffle-seed sensitivity.}
\Cref{fig:fixitv2-seed-repeats} reports two additional shuffle seeds for the fixed 158-record 4ops s1 recipe. Absolute correctness rates and peak speedups vary across seeds, but both repeats produce correct kernels on every d128 training problem, GEMM, all four d64 problems, and both d96 forward problems; neither solves either d96 backward problem. Both produce correct GQA candidates, but none of those candidates execute the target instructions. Together with the seed-0 result in~\Cref{fig:fixitv2-generalization}, these repeats show that the qualitative transfer pattern is stable despite substantial per-problem variation.

\begin{table*}[htbp]
\centering
\scriptsize
\setlength{\tabcolsep}{2pt}
\renewcommand{\arraystretch}{1.12}
\caption{Turn correctness rates for the five-problem SFT evaluation.}
\label{tab:data-recipes-five-problem-turn-success}
\begin{tabular*}{\textwidth}{@{\extracolsep{\fill}}lccccc@{}}
\toprule
SFT-ed Model Label
& GEMM
& MHA-Fwd
& MHA-Fwd-Causal
& MHA-Bwd
& MHA-Bwd-Causal \\
\midrule
Qwen3.6-27B-s1
& \accell{25.0 / 14.6 / 13.5}{25.0 / 14.6 / 13.5}
& \accell{16.7 / 16.7 / 19.8}{16.7 / 16.7 / 19.8}
& \accell{-- / -- / 3.1}{-- / -- / 3.1}
& \accell{-- / 2.1 / 4.2}{-- / 2.1 / 4.2}
& \accell{-- / 2.1 / 5.2}{-- / 2.1 / 4.2} \\

Qwen3.6-27B-s2
& \accell{-- / 2.1 / 2.1}{-- / 2.1 / 2.1}
& \accell{-- / 2.1 / 5.2}{-- / 2.1 / 5.2}
& \accell{-- / -- / 1.0}{-- / -- / 1.0}
& \accell{-- / -- / 1.0}{-- / -- / 1.0}
& -- \\

Qwen3.6-27B-s3
& \accell{-- / 2.1 / 3.1}{-- / 2.1 / 3.1}
& \accell{-- / 4.2 / 2.1}{-- / 4.2 / 2.1}
& \accell{-- / 6.2 / 4.2}{-- / 6.2 / 4.2}
& \accell{-- / 4.2 / 4.2}{-- / 4.2 / 4.2}
& -- \\

Qwen3.6-27B-s4
& \accell{8.3 / 8.3 / 10.4}{8.3 / 8.3 / 10.4}
& \accell{-- / -- / 4.2}{-- / -- / 4.2}
& \accell{-- / 4.2 / 2.1}{-- / 4.2 / 2.1}
& \accell{-- / -- / 1.0}{-- / -- / 1.0}
& -- \\

Qwen3.6-27B-s5
& \accell{16.7 / 14.6 / 12.5}{16.7 / 14.6 / 12.5}
& \accell{-- / 2.1 / 4.2}{-- / 2.1 / 4.2}
& \accell{-- / 2.1 / 3.1}{-- / 2.1 / 3.1}
& \accell{-- / -- / 5.2}{-- / -- / 5.2}
& \accell{-- / -- / 4.2}{-- / -- / 4.2} \\

Qwen3.6-27B-s6
& \accell{8.3 / 2.1 / 1.0}{8.3 / 2.1 / 1.0}
& --
& --
& --
& -- \\
\bottomrule
\end{tabular*}
\end{table*}

\begin{table*}[htbp]
\centering
\scriptsize
\setlength{\tabcolsep}{2pt}
\renewcommand{\arraystretch}{1.12}
\caption{Best speedups for the five-problem SFT evaluation.}
\label{tab:data-recipes-five-problem-speedup}
\begin{tabular*}{\textwidth}{@{\extracolsep{\fill}}lccccc@{}}
\toprule
SFT-ed Model Label
& GEMM
& MHA-Fwd
& MHA-Fwd-Causal
& MHA-Bwd
& MHA-Bwd-Causal \\
\midrule
Qwen3.6-27B-s1
& \spcell{0.303 / 0.303 / 0.340}{0.303 / 0.303 / 0.340}
& \spcell{0.556 / 0.556 / 0.565}{0.556 / 0.556 / 0.565}
& \spcell{-- / -- / \textbf{0.395}}{-- / -- / 0.395}
& \spcell{-- / 0.380 / 0.389}{-- / 0.380 / 0.389}
& \spcell{-- / 0.192 / 0.199}{-- / 0.192 / 0.199} \\

Qwen3.6-27B-s2
& \spcell{-- / 0.209 / 0.211}{-- / 0.209 / 0.211}
& \spcell{-- / 0.548 / 0.548}{-- / 0.548 / 0.548}
& \spcell{-- / -- / 0.315}{-- / -- / 0.315}
& \spcell{-- / -- / 0.296}{-- / -- / 0.296}
& -- \\

Qwen3.6-27B-s3
& \spcell{-- / 0.274 / 0.280}{-- / 0.274 / 0.280}
& \spcell{-- / 0.465 / 0.465}{-- / 0.465 / 0.465}
& \spcell{-- / 0.388 / 0.388}{-- / 0.388 / 0.388}
& \spcell{-- / 0.494 / \textbf{0.494}}{-- / 0.494 / 0.494}
& -- \\

Qwen3.6-27B-s4
& \spcell{0.276 / 0.373 / 0.373}{0.276 / 0.373 / 0.373}
& \spcell{-- / -- / 0.452}{-- / -- / 0.452}
& \spcell{-- / 0.383 / 0.383}{-- / 0.383 / 0.383}
& \spcell{-- / -- / 0.199}{-- / -- / 0.199}
& -- \\

Qwen3.6-27B-s5
& \spcell{0.325 / 0.446 / \textbf{0.446}}{0.325 / 0.446 / 0.446}
& \spcell{-- / 0.425 / \textbf{0.573}}{-- / 0.425 / 0.573}
& \spcell{-- / 0.241 / 0.246}{-- / 0.241 / 0.246}
& \spcell{-- / -- / 0.295}{-- / -- / 0.295}
& \spcell{-- / -- / \textbf{0.246}}{-- / -- / 0.246} \\

Qwen3.6-27B-s6
& \spcell{0.073 / 0.073 / 0.073}{0.073 / 0.073 / 0.073}
& --
& --
& --
& -- \\
\bottomrule
\end{tabular*}
\end{table*}

\begin{table*}[htbp]
\centering
\scriptsize
\setlength{\tabcolsep}{2pt}
\renewcommand{\arraystretch}{1.12}
\caption{Training data recipes.}
\label{tab:data-recipes}
\begin{tabular*}{\textwidth}{@{\extracolsep{\fill}}llcc@{}}
\toprule
SFT-ed Model Label & Config & Reasoning Synthesizer & Record Count \\
\midrule
Qwen3.6-27B-s1 & 4ops & GLM-5.2 & 158 \\
Qwen3.6-27B-s2 & 4ops-Extended & GLM-5.2 & 259 \\
Qwen3.6-27B-s3 & 8ops-Extended & GLM-5.2 & 406 \\
Qwen3.6-27B-s4 & 8ops-Post-balanced & GLM-5.2 & 170 \\
Qwen3.6-27B-s5 & 8ops-Pre-balanced & GLM-5.2 & 258 \\
Qwen3.6-27B-s6 & 8ops-Pre-balanced & Qwen3.6-27B & 258 \\
\bottomrule
\end{tabular*}
\end{table*}

\begin{figure}[H]
\centering
\begin{minipage}{0.98\textwidth}
\begin{lstlisting}[style=prompt]
1. In your thinking, write a short "kernel contract" block listing tensor shapes, flattened offsets, tile sizes, TMA descriptor dimensions/strides/boxDim/swizzle, mbarrier counts, and launch shared-memory bytes. The code must match this block exactly.
2. For TMA BF16 with 128B swizzle and flattened `[B*H*S, D]`, use `globalDim={D, B*H*S}`, `globalStrides={D*2}`, and `boxDim={64,128}` unless the kernel explicitly proves a different layout.
3. Every host-side CUDA/driver call must be checked: `CU_CHECK(cuTensorMapEncodeTiled(...))`, `CUDA_CHECK(cudaFuncSetAttribute(...))`, `CUDA_CHECK(cudaGetLastError())`, and `CUDA_CHECK(cudaStreamSynchronize(stream))`.
4. If dynamic shared memory exceeds the default limit, the kernel must opt in with `cudaFuncSetAttribute` using the same byte count as the launch.
5. The online softmax update must explicitly maintain `m_i`, `l_i`, and rescale old `O` accumulators when the row max changes. Any kernel that computes a tile softmax independently and then accumulates `P @ V` is wrong for full attention.
6. Prohibit unsynchronized shared accumulator updates. Each accumulator element must have a single owner thread/warp, or use an explicit reduction with a documented synchronization point.
7. Require a self-audit checklist at the end of generation: descriptor validity, barrier arrival/tx-count matching, WGMMA descriptor LBO/SBO, output/LSE indexing, and launch shared memory.
\end{lstlisting}
\end{minipage}
\caption{Expert guidance used in MHA evaluation.}
\label{lst:mha-patch}
\end{figure}

\end{document}